\documentclass[letterpaper]{article} 
\usepackage{aaai2026}  
\usepackage{times}  
\usepackage{helvet}  
\usepackage{courier}  
\usepackage[hyphens]{url}  
\usepackage{graphicx} 
\usepackage{natbib}  
\usepackage{caption} 
\usepackage{algorithm}
\usepackage{algorithmic}

\usepackage{newfloat}
\usepackage{listings}
\DeclareCaptionStyle{ruled}{labelfont=normalfont,labelsep=colon,strut=off} 
\floatstyle{ruled}
\newfloat{listing}{tb}{lst}{}
\floatname{listing}{Listing}
\title{Look as You Think: Unifying Reasoning and Visual Evidence Attribution for Verifiable Document RAG via Reinforcement Learning}
\author{
    Shuochen Liu,
    Pengfei Luo,
    Chao Zhang,
    Yuhao Chen,
    Haotian Zhang,
    Qi Liu,
    Xin Kou, \\
    Tong Xu\thanks{Corresponding author.},
    Enhong Chen
}
\affiliations{
    University of Science and Technology of China\\

    shuochenliu@mail.ustc.edu.cn, tongxu@ustc.edu.cn
}

\usepackage{bibentry}

\usepackage{amsmath} 
\usepackage{amssymb}
\usepackage{booktabs} 
\usepackage{multirow}
\usepackage{subcaption}

\begin{document}

\maketitle

\begin{abstract}
Aiming to identify precise evidence sources from visual documents, visual evidence attribution for visual document retrieval–augmented generation (VD-RAG) ensures reliable and verifiable predictions from vision-language models (VLMs) in multimodal question answering. Most existing methods adopt end-to-end training to facilitate intuitive answer verification. However, they lack fine-grained supervision and progressive traceability throughout the reasoning process. In this paper, we introduce the \textbf{Chain-of-Evidence (CoE)} paradigm for VD-RAG. CoE unifies Chain-of-Thought (CoT) reasoning and visual evidence attribution by grounding reference elements in reasoning steps to specific regions with bounding boxes and page indexes. To enable VLMs to generate such evidence-grounded reasoning, we propose \textbf{Look As You Think (LAT)}, a reinforcement learning framework that trains models to produce verifiable reasoning paths with consistent attribution. During training, LAT evaluates the attribution consistency of each evidence region and provides rewards only when the CoE trajectory yields correct answers, encouraging process-level self-verification. Experiments on vanilla Qwen2.5-VL-7B-Instruct with Paper‑ and Wiki‑VISA benchmarks show that LAT consistently improves the vanilla model in both single- and multi-image settings, yielding average gains of 8.23\% in soft exact match (EM) and 47.0\% in IoU@0.5. Meanwhile, LAT not only outperforms the supervised fine-tuning baseline, which is trained to directly produce answers with attribution, but also exhibits stronger generalization across domains.
\end{abstract}

\section{Introduction}
With the development of multimodal understanding capabilities in vision-language models (VLMs)~\cite{chen2025expandingperformanceboundariesopensource, bai2025qwen25vltechnicalreport}, visual document retrieval-augmented generation (VD-RAG) has emerged as a critical research frontier. Nevertheless, current VLMs remain susceptible to hallucinations, whereby their outputs may deviate from the source document content~\cite{bai2025hallucinationmultimodallargelanguage}. Without reliable visual evidence attribution mechanisms to identify sources in documents, users cannot intuitively trace back the specific information employed by the model, thereby reducing the reliability of VD-RAG systems in applications. 

\begin{figure}[t] 
\centerline{
\includegraphics[width=0.97\linewidth,scale=1.00]{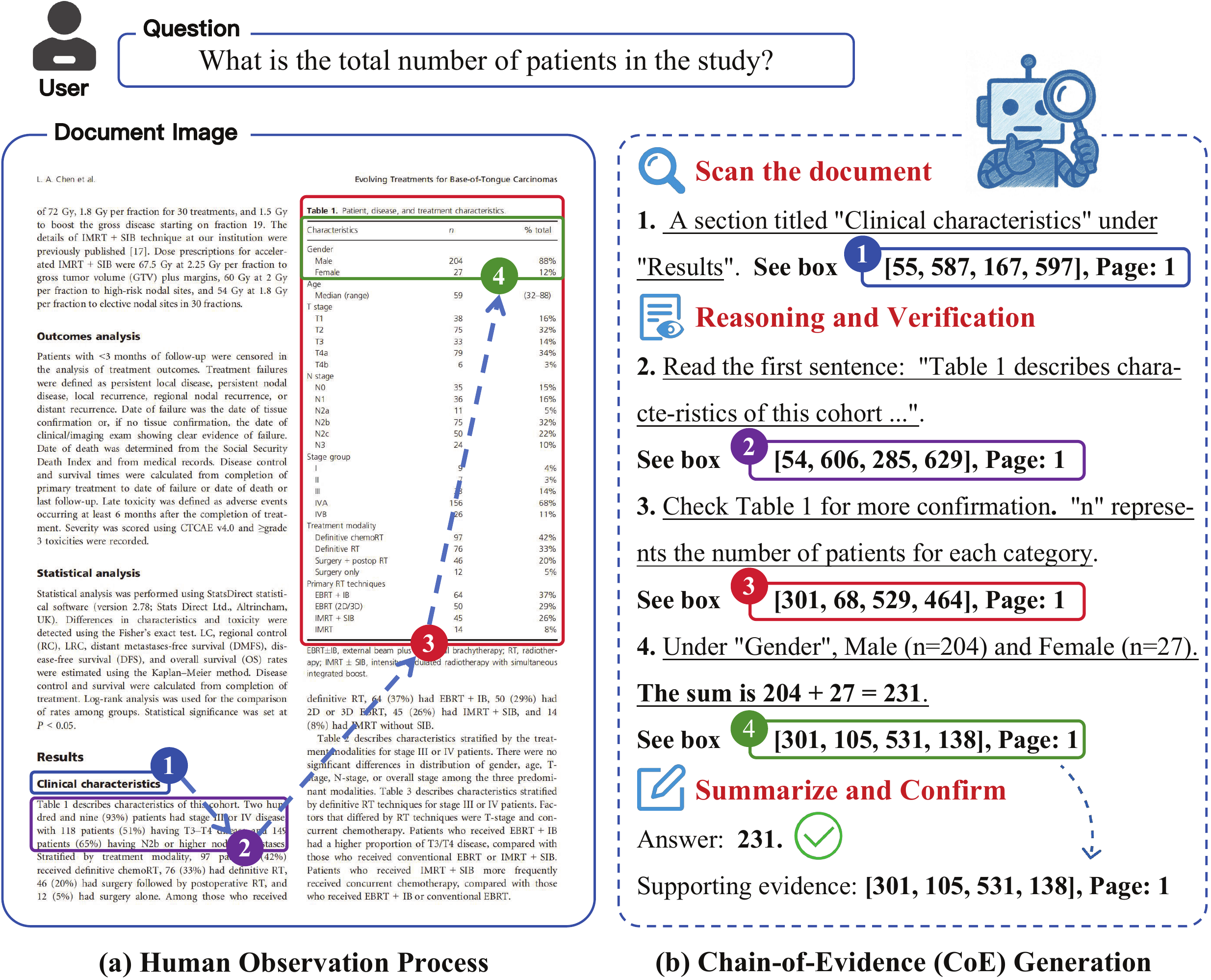} } 
\caption{ 
(a) Humans infer information by observing and locating supporting evidence in the document. (b) Each element in the reasoning step is linked to a visual attribution via a bounding box during Chain-of-Evidence generation.
} 
\label{fig:overview} 
\end{figure}

Along this line, recent works~\cite{shao2024visual, wu2025groundedchainofthoughtmultimodallarge, qi2025cogcomvisuallanguagemodel} mitigate hallucination in single-image Chain-of-Thought (CoT)~\cite{kojima2022large} reasoning by attending to critical visual regions as evidence, leveraging large-scale annotations of stepwise evidence regions. 
VISA~\cite{ma2024visaretrievalaugmentedgeneration} further incorporates visual evidence attribution into the VD-RAG framework by linking answers with supporting sources via bounding boxes. Despite these advances, effective evidence attribution in visual documents still faces challenges: \textbf{(1)~Lack of progressive reasoning mechanisms for verifiable attribution.} Existing methods, such as VISA, directly associate answers with final evidence, but do not reveal the intermediate process that would help users clearly understand the content and trace how the evidence is located. In contrast, when addressing complex problems, humans do not directly locate evidence but progressively search for relevant information. As shown in Figure~\ref{fig:overview}a, such an observation pathway is structured as ``chapter-paragraph summary-specific element'', localizing evidence from coarse to fine. However, current VLMs struggle to replicate this progressive observation process, especially in multi-image scenarios.
\textbf{(2)~Limited supervision for learning multimodal reasoning.} Training models to learn CoT reasoning requires extensive annotated data~\cite{xu2025llavacotletvisionlanguage, wang2025vgrvisualgroundedreasoning}, especially for stepwise evidence attribution. However, manually constructing such datasets is a costly process. Therefore, how to effectively learn stepwise visual evidence attribution and generalize reasoning abilities under limited annotated data remains a critical challenge.

Given these challenges, we introduce \textbf{Chain-of-Evidence (CoE)}, a reasoning paradigm that integrates Chain-of-Thought (CoT) reasoning with visual evidence attribution, designed for VD-RAG. Unlike existing methods that perform reasoning within a single image or directly link the final answer to the source, CoE models the intermediate reasoning trajectory by grounding each step to a supporting source from documents. This coarse-to-fine attribution process mirrors human problem-solving and enhances the reliability of VD-RAG. To this end, we introduce \textbf{\underline{L}ook-\underline{A}s-You-\underline{T}hink (LAT)}, a two-stage training approach designed to implement the proposed CoE paradigm. Specifically, in the first stage, we fine-tune VLM on a set of few human-verified annotated CoE data to learn reasoning patterns. Then we adopt reinforcement learning (RL) under the Group Relative Policy Optimization (GRPO)~\cite{shao2024deepseekmathpushinglimitsmathematical} through a tailored reward design. The model is guided by a stepwise reward based on the semantic alignment between predicted visual evidence and corresponding context, encouraging faithful reasoning without requiring stepwise annotations. Furthermore, the reward is issued only when the CoE trajectory yields the correct answer. We further incorporate the outcome reward to ensure answer accuracy and guide the model in defining the scope of answer localization. This combined reward scheme enhances the ability to generate CoE reasoning, thereby linking sub-step verification to end-task performance. Our contributions are summarized as follows:

\noindent\textbf{1}) In the VD-RAG scenario, we formalize the \textbf{Chain-of-Evidence (CoE)} paradigm by modeling multimodal reasoning as a sequence of grounded steps, where reference elements (e.g., figures, tables, or factual information) are linked to their source through a bounding box and page index.

\noindent\textbf{2}) Building upon CoE, we propose \textbf{LAT}, an RL-based approach that jointly optimizes reasoning and visual grounding through a stepwise reward. By aligning each reference element with its visual evidence, LAT enables attribution-aware reasoning with few CoE-annotated samples.

\noindent\textbf{3}) LAT balances traceability and performance. Compared to the vanilla model, it shows average improvements of 8.23\% EM and 47.0\% IoU@0.5 in both single- and multi-image scenarios. Meanwhile, LAT not only outperforms the supervised fine-tuning baseline, which is trained to directly produce answers with attribution without CoE reasoning, but also exhibits stronger cross-domain generalization.

\section{Related Work}

\subsection{Visual Evidence Attribution}
Early end-to-end grounding methods integrate object detection into generated text by using markdown hyperlinks to generate bounding-box tokens~\cite{chen2023shikraunleashingmultimodalllms, peng2023kosmos2groundingmultimodallarge}. They are trained on large corpora of grounded images and texts. Building on this foundation, multi-step visually grounded CoT frameworks~\cite{shao2024visual, li2025vocot, wu2025groundedchainofthoughtmultimodallarge, xia2025bootstrappinggroundedchainofthoughtmultimodal} interleave reasoning and localization by predicting bounding boxes as evidence within the reasoning process, thereby yielding interpretable reasoning traces. However, these approaches have been validated exclusively on general visual perception tasks and rely on large-scale annotated evidence regions. They also do not explore the visual evidence attribution task in VD-RAG involving heterogeneous layouts and multi-page retrieval.

Moreover, existing text-based evidence attribution methods~\cite{gao2023enabling, ye2024effective} in document RAG often operate at the document level, requiring users to read entire documents to locate supportive content. VISA \cite{ma2024visaretrievalaugmentedgeneration} first adapts visual evidence attribution to document screenshots by aligning the final answer with its evidence. Despite enabling intuitive verification of correctness, it fails to explicate the intermediate reasoning steps through which the model arrives at the answer. These limitations motivate the CoE reasoning paradigm, which generates faithful reasoning steps and validates each reference element against its corresponding source, as shown in Figure~\ref{fig:overview}b.

\subsection{RL for VLM Reasoning}
Recent research has shown that RL-based policy optimization~\cite{zhang2024item} can improve the reasoning capabilities of large language models (LLMs)~\cite{chen2025xiangqir1enhancingspatialstrategic, chen-etal-2025-think, zhang2025tearagtokenefficientagenticretrievalaugmented}. DeepSeek-R1~\cite{deepseekai2025deepseekr1incentivizingreasoningcapability} demonstrated that RL training elicits emergent CoT behaviors in LLMs, revealing the hallmark ``aha moment''. Inspired by this phenomenon, several methods~\cite{peng2025lmmr1empowering3blmms} have extended R1-style RL strategies to VLMs, leveraging rule-based reward functions to boost performance on mathematical reasoning and visual perception tasks. 
Unlike supervised fine-tuning (SFT), RL-based approaches achieve deeper reasoning and stronger generalization without relying on extensive human-annotated data~\cite{chu2025sftmemorizesrlgeneralizes}.

However, existing RL frameworks for multimodal tasks are primarily optimized for answer accuracy as the reward signal~\cite{yang2025r1onevisionadvancinggeneralizedmultimodal, shen2025vlmr1stablegeneralizabler1style}, with no explicit supervision for verifying intermediate reasoning~\cite{ni2025pointrftimprovingmultimodalreasoning, cao2025groundr1incentivizinggroundedvisual}, and without design considerations for the visual evidence attribution task in VD-RAG. Drawing on human reading strategies, we introduce a stepwise, process-level reward that aligns each reasoning step with verifiable evidence. Leveraging the extracted CoE, we explicitly reward trajectories that are evidentially consistent and culminate in the correct answer, thereby ensuring valid reasoning and fostering faithful, attribution-based explanations.

\begin{figure*}[t] 
\centerline{
\includegraphics[width=0.97\textwidth]{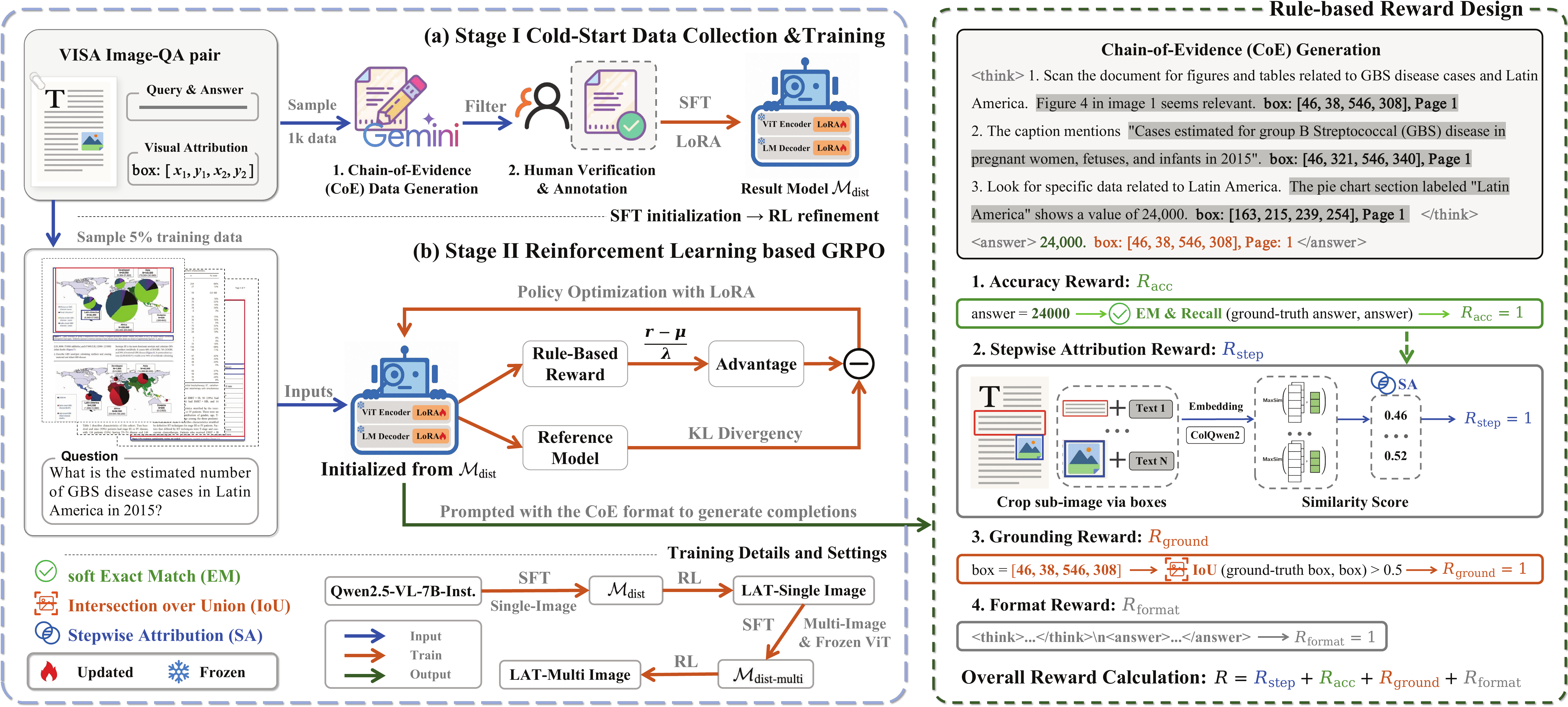} } 
\caption{ 
Overview of the proposed LAT framework.
Left: A two-stage training pipeline. Stage I generates and filters the CoE data for fine-tuning. Stage II: The model undergoes refinement via RL under the GRPO algorithm. Right: Rule-based reward design. In GRPO training, the model generates CoE reasoning to guide policy updates through the reward signals.
} 
\label{fig:framework} 
\end{figure*}

\section{Proposed Approach}
\label{sec:Proposed Approach}
To remedy the lack of verifiable progressive reasoning for visual evidence attribution, we first formalize the \textbf{Chain-of-Evidence (CoE)} paradigm. Building upon CoE, we present \textbf{LAT}, a two-stage RL-based framework shown in Figure~\ref{fig:framework}. Stage I performs supervised fine-tuning to align annotated CoE traces, and Stage II conducts fine-grained reinforcement learning with an answer-conditioned attribution reward that refines step-level grounding. 

\subsection{CoE Formalization and Notations}
We formalize the generation with Chain-of-Evidence (CoE) reasoning as follows. Specifically, we define a textual query $q$ and a set of document pages $\mathcal{P}=\{p_n\}_{n=1}^N$, which are pre-retrieved from the corpus. Given $(q, \mathcal{P})$, CoE requires a VLM $\phi$ to perform CoT reasoning with stepwise visual attribution and then produce both an answer and its supporting evidence for evaluation, formulated as:
\begin{equation}
\mathcal{R}, \mathcal{B}, \mathcal{A}= \phi(q, \mathcal{P}).
\end{equation}

Here, $\mathcal{R}=\{r_t\}_{t=1}^T$ denotes the textual reasoning steps, while $\mathcal{B} = \{(i_t, B_t)\}_{t=1}^T$ is the corresponding evidence chain, where $i_t\in[1, N]$ indicates the page index and $B_t=[(x_1^t, y_1^t), (x_2^t, y_2^t)]$ specifies the bounding box of the visual evidence for $r_t$ on $i_t$-th page. After CoE reasoning, the final output $\mathcal{A} = \{a, (i^*, B_{ans})\}$ consists of the answer $a$, the most relevant page $p_{i^*}\in \mathcal{P}$, and the bounding box $B_{ans} \in \mathcal{B}$ confirming the evidence that supports $a$. It should be noted that not all reasoning steps in the response require a bounding box over images, such as calculation or conclusion steps. We omit these cases in our formulation for mathematical conciseness.

\subsection{Stage I: Cold-Start Data Collection and Training}
\label{cold}
To prime the model for CoE reasoning, we first sample $1,000$ instances from each training dataset and prompt a stronger proprietary model Gemini 2.5 Pro~\cite{comanici2025gemini25pushingfrontier} using two CoE exemplars as in-context demonstrations. The model produces stepwise rationales, each annotated with bounding-box evidence, yielding a cold-start corpus that reflects the desired reasoning format. 

For each query $q$ and its corresponding response $y$, we assess answer quality with a recall metric,
\begin{equation}
\text{Recall}(a, a_{gt})\;=\;
\frac{\lvert a\,\cap\,a_{gt}\rvert}{\lvert a_{gt}\rvert}, \label{eq:recall}
\end{equation}
where $a_{gt}$ is the ground-truth answer from the dataset, $\lvert a \rvert$ denotes the number of words in the extracted answer portion $a$ from the response $y$, and $\lvert a\,\cap\,a_{gt}\rvert$ is the number of overlapping words between $a$ and $a_{gt}$. Only samples with recall above a threshold $\gamma$ are retained to ensure sufficient answer accuracy in the initial CoE traces.

We manually verify and correct any bounding-box drift, retaining verified samples with correct answers ($\sim$30\%) to ensure data quality. Details of the resulting dataset $\mathcal{D}_{\text{final}}$ used in the cold-start training, including its split distribution, are provided in Table~\ref{tab:visa-dataset-stats} in the appendix. Next, we fine-tune the VLM on $\mathcal{D}_{\text{final}}$ using LoRA~\cite{hu2022lora}, aiming to minimize the cross-entropy loss between the generated output and the annotated reasoning sequences through SFT. The resulting model is defined as $\mathcal{M}_\text{dist}$.

\subsection{Stage II: Unified Reasoning and Visual Attribution via Reinforcement Learning}
\label{reward_design}

To emulate the human observation process shown in Figure~\ref{fig:overview}, the model needs to deliver both an accurate answer and attribution-aware CoE reasoning to identify evidence supporting the final answer. We decompose the overall objective into four sub-goals: answer accuracy, stepwise visual attribution quality, evidence grounding precision, and adherence to the structured output format. Accordingly, we design four reward functions for rule-based RL training.

\subsubsection{Accuracy Reward ($R_{\mathrm{acc}}$).}
We reward the model based on soft exact match (EM), considering a response correct and $\mathrm{EM}(a, a_{gt})$ equal to 1 if the normalized predicted answer $a$ is a substring of the ground truth $a_{gt}$, or vice versa. To prevent the reward from becoming too sparse, we enhance this signal by including the recall metric. Formally,
\begin{align}
R_{\mathrm{acc}} = \frac{\mathbb{I}\left(\mathrm{EM}(a, a_{gt}) = 1\right) + \text{Recall}(a, a_{gt})}
       {2}, \label{eq:acc}
\end{align}
where $\mathbb{I}(\cdot)$ denotes the indicator function. By incorporating the recall metric, we prevent the wholesale rejection of semantically relevant yet not exact-match samples, while assigning higher rewards to perfectly matched outputs, thereby enhancing the model’s generalization capability.

\subsubsection{Stepwise Attribution Reward ($R_{\mathrm{step}}$).}
To ensure each reasoning step is grounded in the associated evidence attribution, we design a stepwise reward that measures the semantic alignment between each step and the evidence indicated by the bounding box.
For each textual reasoning $r_t$ involving reference element, we crop the raw document image of the $i_t$-th page according to the predicted bounding box $B_t$ and encode both the cropped sub-image and $r_t$ using a multimodal retriever ColQwen2~\cite{faysse2024colpali}, which enables efficient document indexing through visual features and handles dynamic resolutions. Formally,
\begin{align}
e^{(t)}_{\mathrm{img}} &= \mathrm{Norm}(\mathrm{enc}_{\mathrm{img}}(\mathrm{\textbf{crop}}(B_t))),  \\
e^{(t)}_{\mathrm{txt}} &= \mathrm{Norm}(\mathrm{enc}_{\mathrm{txt}}(r_t)),
\end{align}
where $\mathrm{Norm}(\cdot)$ denotes $\text{L}_2$ normalization for consistent cosine similarity evaluation. $\mathrm{enc}_{\mathrm{img}}(\cdot)$ and $\mathrm{enc}_{\mathrm{txt}}(\cdot)$ represent the image and text encoders of the retriever, respectively. $\mathrm{\textbf{crop}}(\cdot)$ denotes the extraction of a sub‐image from the original document page, delimited by the bounding box $B_t$. The stepwise attribution reward is defined as:
\begin{align}
S &= \min_{1 \le k \le K} \cos(e_{\mathrm{img}}^{(k)},\, e_{\mathrm{txt}}^{(k)}), \\
I &= \max_{1 \le i,j \le K,\; i \neq j} \text{IoU}(B_i, B_j), \label{eq:IoU_formula} \\
R_{\mathrm{step}} &= \frac{\mathbb{I}(S \ge \tau)
       + \mathbb{I}(I \le \delta)}
       {2}\,\cdot \mathbb{I}({R_{\mathrm{acc}} \ge \epsilon}), \label{eq:threshold}
\end{align}
where $K$ is the number of context-evidence pairs and $\cos(\cdot)$ denotes the cosine similarity measure. We require all pairs in the CoE reasoning process to exceed the similarity threshold $\tau$ to ensure proper alignment. Nevertheless, the model may exploit this mechanism by repeating bounding boxes to satisfy the reward, which undermines coarse-to-fine evidence attribution and rich visual cues.

To mitigate such a phenomenon, we introduce a constraint on bounding box overlap by computing the maximum pairwise intersection over union (IoU) $I$ among bounding boxes from the reasoning steps (Equation~\ref{eq:IoU_formula}) and enforcing $I \le \delta$ in Equation~\ref{eq:threshold}. This constraint promotes attribution diversity and encourages progressive grounding across reasoning steps. Furthermore, the reward is constrained by $R_{\mathrm{acc}}$, ensuring that only faithful CoE processes contributing to correct answers are reinforced. This formulation guides the model to treat visual grounding as an internal retrieval problem, supporting stepwise and fine-grained visual attribution.

\subsubsection{Grounding Reward ($R_{\mathrm{ground}}$).}
The model needs to select the relevant source from the CoE reasoning process as evidence to support the final answer. We measure its precision by computing the IoU between the predicted bounding box $(i^*, B_{ans})$ and the ground-truth evidence $(i_{gt}, B_{gt})$:
\begin{equation}
R_{\mathrm{ground}} = \mathbb{I}\left(\text{IoU}(B_{ans}, B_{gt}) > 0.5\right) \label{ground}, \text{s.t. } i^* = i_{gt}.
\end{equation}

By penalizing misaligned evidence attribution, $R_{\mathrm{ground}}$ drives the model to identify correct content within the source page $i_{gt}$ instead of blindly collecting irrelevant information.

\subsubsection{Format Reward ($R_{\mathrm{format}}$).} We prompt the model to conduct CoE reasoning on visual documents and generate answers. Well-formatted outputs contain CoE reasoning within \texttt{<think>...</think>} tags and answers with supporting evidence within \texttt{<answer>...</answer>} tags.
\begin{equation}
R_{\text{format}} = 
\begin{cases}
1, & \text{if the format is correct}, \\
-1, & \text{otherwise}.
\end{cases}
\label{eq:format}
\end{equation}

By assigning negative rewards to incorrectly formatted outputs, we impose stricter output constraints and accelerate convergence toward the desired format, enabling more targeted optimization of the remaining reward components.

\subsubsection{Training Algorithm.}
\label{RL_grpo}
As shown in Figure~\ref{fig:framework}, we sample 5\% training data from the raw dataset, initialize the policy model $\pi_\theta$ from the cold-start checkpoint $\mathcal{M}_\text{dist}$, and adopt the GRPO~\cite{shao2024deepseekmathpushinglimitsmathematical} algorithm, combined with the fine-grained reward design~\cite{shen2025vlmr1stablegeneralizabler1style}. GRPO, which supports rule-based rewards and optimizes the policy $\pi_\theta$ by sampling a group of candidate outputs for each query and computing a group-relative advantage, eliminates the need for training a critic model. We iteratively update $\pi_\theta$, encouraging it to produce responses with higher relative advantages within the sampled group under the GRPO objective.

\begin{table*}[t]
\centering
\small
\begin{tabular}{l|c|cc|cc|cc|cc}
\toprule
\textbf{Models} & \textbf{Param.} & \multicolumn{2}{c|}{\textbf{Wiki-VISA (Single)}} & \multicolumn{2}{c|}{\textbf{Paper-VISA (Single)}} & \multicolumn{2}{c|}{\textbf{Wiki-VISA (Multi)}} & \multicolumn{2}{c}{\textbf{Paper-VISA (Multi)}} \\
\cmidrule(lr){3-4} \cmidrule(lr){5-6} \cmidrule(lr){7-8} \cmidrule(lr){9-10}
& \textbf{Size} & EM & IoU@0.5 & EM & IoU@0.5 & EM & IoU@0.5 & EM & IoU@0.5 \\
\midrule
\multicolumn{1}{c|}{} & \multicolumn{1}{c|}{} & \multicolumn{8}{c}{\textbf{Proprietary and Open-source Models, Direct Answer}} \\
Gemini2.0 flash \dag & - & 73.3 & - & 46.8 & - & 28.2 & - & 26.7 & -\\
Qwen-VL max \dag & - & 63.1 & - & 46.6 & - & 48.7 & - & 32.1 & -\\
mPLUG-DocOwl2 & 8B & 23.6 & - & 20.3 & - & 19.3 & - & 21.5 & - \\
Qwen2.5-VL & 7B & 67.7 & - & 38.2 & - & 54.1 & - & 34.6 & - \\
Qwen2.5-VL & 32B & 66.3 & - & 36.0 & - & 60.8 & - & 38.8 & - \\
InternVL2.5 & 8B & 45.9 & - & 42.4 & - & 37.7 & - & 32.1 & -\\
LLaVA-OneVision & 7B & 48.4 & - & 28.7 & - & 38.5 & - & 30.8 & - \\
LLaVA-CoT & 11B & 45.3 & - & 28.8 & - & - & - & - & - \\
R1-OneVision & 7B & 58.0 & - & 24.4 & - & 57.2 & - & 31.4 & -\\
\midrule
\multicolumn{1}{c|}{} & \multicolumn{1}{c|}{} & \multicolumn{8}{c}{\textbf{Proprietary Model, Full Data Training, Direct Answer \& Attribution}} \\
VISA \dag & 7B & 74.1 & 28.3 & 49.6 & 62.2 & 58.7 & 31.6 & 54.1 & 65.6 \\
\midrule
Qwen2.5-VL (DA) & 7B & 69.4 & 0.80 & 43.8 & 2.87 & 52.5 & 0.97 & 37.2 & 5.56 \\
\textbf{LAT-Ind.} & 7B & \textbf{73.6} & 53.7 & 45.4 & \textbf{49.9} & 64.5 & 38.0 & \textbf{51.2} & 46.9 \\
\textbf{LAT-Full} & 7B & 73.1 & \textbf{57.8} & \textbf{46.2} & 48.4 & \textbf{64.8} & \textbf{41.4} & 50.6 & \textbf{49.3}\\
$\Delta$ Vanilla model & - & +4.20 & +57.0 & +2.40 & +47.0 & +12.3 & +40.4 & +14.0 & +43.7\\
\bottomrule
\end{tabular}
\caption{Performance comparison on Paper- and Wiki-VISA in both single- and multi-image settings. Bold indicates the best score in each column. $\dag$ denotes the proprietary model, which is fine-tuned on the full in-domain dataset, serving as an upper-bound baseline. ``DA'' refers to the zero-shot prompting setting for direct answer \& attribution without training.}
\label{tab:direct_answer_result}
\end{table*}

\section{Experiment Setup}

\subsection{Datasets}
\label{dataset}
We conducted the experiments on the VISA benchmark~\cite{ma2024visaretrievalaugmentedgeneration}, which is the first dataset designed for visual evidence attribution in real-world VD-RAG scenarios and comprises three subsets:
(1)~\textbf{Wiki-VISA} is derived from the Natural Questions (NQ)~\cite{kwiatkowski2019natural}. VISA renders the Wikipedia pages and identifies the HTML element containing the answer with a bounding box. (2)~\textbf{Paper-VISA} builds upon PubLayNet~\cite{zhong2019publaynet}. VISA synthesizes QA pairs grounded in annotated layouts. (3)~\textbf{FineWeb-VISA} extends FineWeb-edu~\cite{penedo2024fineweb} by selecting passages longer than 50 tokens and synthesizing grounded QA pairs. In the single-image setup, each query is paired with a source document page, a short answer extracted from it, and an evidence bounding box.

When extended to multiple images, for each query $q$, two additional document images are randomly sampled from the top $K$ screenshots retrieved by Document Screenshot Embedding (DSE)~\cite{ma-etal-2024-unifying} within the VISA dataset, then merged with the source document page as input. For no-answer scenarios, the source document page is replaced with an irrelevant page from the dataset. Example cases are shown in Figure~\ref{fig:visa} in Appendix~\ref{data col}.

\subsection{Baseline}
To evaluate the effectiveness of LAT, we compared it against three categories of baselines: \textbf{(1)~Proprietary and open-source models}, including Gemini~\cite{geminiteam2025geminifamilyhighlycapable}, Qwen-VL Max~\cite{bai2023qwenvlversatilevisionlanguagemodel}, mPLUG-DocOwl2~\cite{hu2024mplugdocowl2highresolutioncompressingocrfree}, InternVL2.5~\cite{chen2025expandingperformanceboundariesopensource}, LLaVA-OneVision~\cite{li2024llavaonevisioneasyvisualtask}, and Qwen2.5-VL~\cite{bai2025qwen25vltechnicalreport}. \textbf{(2)~Reasoning models}, including LLaVA-CoT~\cite{xu2025llavacotletvisionlanguage}, which is trained via SFT on CoT data, and R1-OneVision~\cite{yang2025r1onevisionadvancinggeneralizedmultimodal}, which is optimized with RL in general scenarios. \textbf{(3)~Attribution-supervised model:} VISA-7B~\cite{ma2024visaretrievalaugmentedgeneration}, trained to generate answers with direct attribution. These models serve as baselines to evaluate whether LAT balances answer accuracy with attribution and achieves CoE reasoning under limited supervision.

\subsection{Training Details}
We used Qwen2.5-VL-7B-Instruct~\cite{bai2025qwen25vltechnicalreport} as the backbone and applied LoRA~\cite{hu2022lora} for parameter-efficient fine-tuning, with rank $r{=}64$ and scaling factor $\alpha{=}64$. Following the pipeline in Figure~\ref{fig:framework}, we performed SFT on $\mathcal{D}_{\text{final}}$ using a learning rate of $1\mathrm{e}{-4}$, followed by RL with $5\mathrm{e}{-5}$. During RL training, LAT was trained on 5\% of QA pairs sampled from the raw dataset. After each stage, we merged the LoRA parameters for subsequent training.

In the multi-image setting, each query is paired with three retrieved documents provided in the \texttt{candidates} field, as included in the VISA dataset. When no relevant information is available, the model is trained to output ``No answer''. We initialized the multi-image model from the single-image trained version and further performed SFT using multi-image CoE data in $\mathcal{D}_{\text{final}}$, fine-tuning the LoRA adapter of the LM while keeping the vision transformer (ViT) frozen to reduce GPU memory usage. Additional experimental details are provided in Appendix~A. Our code is available at https://github.com/PolarisLiu1/LAT.

\begin{table*}[t]
\centering
\small
\setlength{\tabcolsep}{1mm}
\begin{tabular}{l|c|ccc|ccc|ccc|ccc}
\toprule
\textbf{Models} & \textbf{Param.} & \multicolumn{3}{c|}{\textbf{Wiki-VISA (Single)}} & \multicolumn{3}{c|}{\textbf{Paper-VISA (Single)}} & \multicolumn{3}{c|}{\textbf{Wiki-VISA (Multi)}} & \multicolumn{3}{c}{\textbf{Paper-VISA (Multi)}}\\
\cmidrule(lr){3-5} \cmidrule(lr){6-8} \cmidrule(lr){9-11} \cmidrule(lr){12-14}
& \textbf{Size} & EM & IoU@0.5 & SA & EM & IoU@0.5 & SA & EM & IoU@0.5 & SA & EM & IoU@0.5 & SA \\
\midrule
\multicolumn{1}{c|}{} & \multicolumn{1}{c|}{} & \multicolumn{12}{c}{\textbf{Proprietary Models, CoE Reasoning}} \\
Gemini2.0 flash \dag & - & 52.0 & 4.47 & 24.0 & 46.4 & 4.72 & 14.1 & 32.1 & 1.20 & 5.20 & 27.0 & 3.24 & 20.0 \\
Qwen-VL max \dag & - & 55.2 & 0.17 & 10.3 & 47.8 & 2.87 & 8.6 & 35.8 & 0.10 & 5.90 & 32.6 & 1.85 & 12.5\\
\midrule
\multicolumn{1}{c|}{} &  \multicolumn{1}{c|}{} & \multicolumn{12}{c}{\textbf{Open-source Models, CoE Reasoning}} \\
Qwen2.5-VL & 7B & 60.4 & 2.20 & 13.9 & 36.9 & 3.80 & 12.4 & 54.9 & 11.0 & 12.5 & 39.6 & 16.2 & 29.5\\
\ \ +one shot ICL & 7B & 61.8 & 1.37 & 12.2 & 37.3 & 8.30 & 29.3 & 54.2 & 9.20 & 11.9 & 39.5 & 18.7 & 30.9 \\
Qwen2.5-VL & 32B & 62.8 & 9.27 & 0.16 & 35.2 & 2.50 & 0.11 & 62.5 & 18.7 & 0.60 & 43.3 & 19.0 & 0.40\\
\ \ +one shot ICL & 32B & 61.7 & 9.13 & 5.23 & 35.0 & 4.12 & 1.25 & 64.1 & 22.1 & 1.87 & 42.6 & 19.6 & 0.51\\
\midrule
\textbf{LAT-Ind.} & 7B & \textbf{73.6} & 53.7 & \textbf{64.6} & 45.4 & \textbf{49.9} & \textbf{35.5} & 64.5 & 38.0 & 71.8 & \textbf{51.2} & 46.9 & 46.3\\
\textbf{LAT-Full} & 7B & 73.1 & \textbf{57.8} & 59.6 & \textbf{46.2} & 48.4 & 33.8 & \textbf{64.8} & \textbf{41.4} & \textbf{75.2} & 50.6 & \textbf{49.3} & \textbf{53.2} \\
$\Delta$ Vanilla model & - & +13.2 & +55.6 & +50.7 & +9.3 & +46.1 & +23.1 & +9.9 & +30.4 & +62.7 & +11.6 & +33.1 & +23.7\\
\bottomrule
\end{tabular}
\caption{Performance comparison for Chain-of-Evidence (CoE) reasoning processes on Paper- and Wiki-VISA datasets. Bold indicates the best score in each column. Results include both result accuracy (EM, IoU@0.5) and process quality (SA) metrics.}
\label{tab:coe_reasoning_result}
\end{table*}

\subsection{Evaluation}

We measured the performance across three dimensions: answer accuracy, evidence grounding, and stepwise attribution quality. Specifically, we reported answer accuracy using \textbf{soft Exact Match (EM)}, and evaluated grounding precision by computing \textbf{IoU@0.5}, which measures the proportion of the predicted box $B_{ans}$ whose IoU with the ground truth evidence exceeds 0.5. To evaluate the quality of stepwise visual attribution, we employed the \textbf{Stepwise Attribution (SA)} reward function. We adopted the default threshold $\tau{=}0.3$ in Equation~\ref{eq:threshold} to determine whether a step is correctly verified.

Following the evaluation of VISA~\cite{ma2024visaretrievalaugmentedgeneration}, we assessed the performance on both the Paper- and Wiki-VISA datasets under two settings. (1)~\textbf{Single-image}: The model is provided solely with the source document and evaluated across three dimensions. (2)~\textbf{Multi-image}: The model is additionally required to identify the source document from a set of retrieved candidates. For reproducibility, we adopt greedy decoding as the decoding strategy during evaluation.

\section{Results and Analysis}
\subsection{Main Results}

\subsubsection{Attribution-aware performance under in-domain and cross-domain settings.} To assess both answer accuracy and attribution precision, we evaluated models under direct answer and attribution-aware settings. We define an in-domain setup as evaluation on datasets from the same distribution as the training data. As shown in Table~\ref{tab:direct_answer_result}, LAT trained on in-domain data (\textbf{LAT-Ind.}) outperforms open-source models. 
Compared to the vanilla model, LAT enhances both answer correctness (EM, \textbf{+7.95\%}) and evidence grounding (IoU@0.5, \textbf{+44.6\%}) by optimizing attribution-aware reasoning through CoE-guided RL. The improvements are evident in multi-page scenarios, where the complexity of evidence selection emphasizes the advantages of our approach. In particular, for unanswerable cases, LAT outputs ``No answer'' and achieves an average precision of 65\%, highlighting its robustness in handling such scenarios. 

Table~\ref{tab:coe_reasoning_result} highlights the effectiveness of LAT in improving CoE reasoning quality. To assess the impact of in-context learning (ICL)~\cite{brown2020language}, we included an annotated CoE example as a prompt. While this yields a moderate improvement (SA, +4.0\%), suggesting that demonstrations can partially guide the generation of structured reasoning, it has a limited effect on the result. LAT achieves a substantial SA gain of \textbf{37.5\%}, demonstrating its ability to ground each reasoning step accurately. Moreover, LAT maintains high answer accuracy, indicating an alignment between faithful reasoning and correct outcomes. To assess generalization across domains, we trained \textbf{LAT-Full} on the sampled subset from all datasets. Compared to the in-domain variant, \textbf{LAT-Full} shows further improvements (e.g., Wiki-VISA Multi, IoU@0.5: 38.0\%$\rightarrow$\textbf{41.4\%}; SA: 71.8\%$\rightarrow$\textbf{75.2\%}), exhibiting generalization across diverse document distributions.

\begin{figure*}[t]
    \centering
    \begin{subfigure}[b]{0.32\textwidth}
        \centering
        \includegraphics[width=\textwidth]{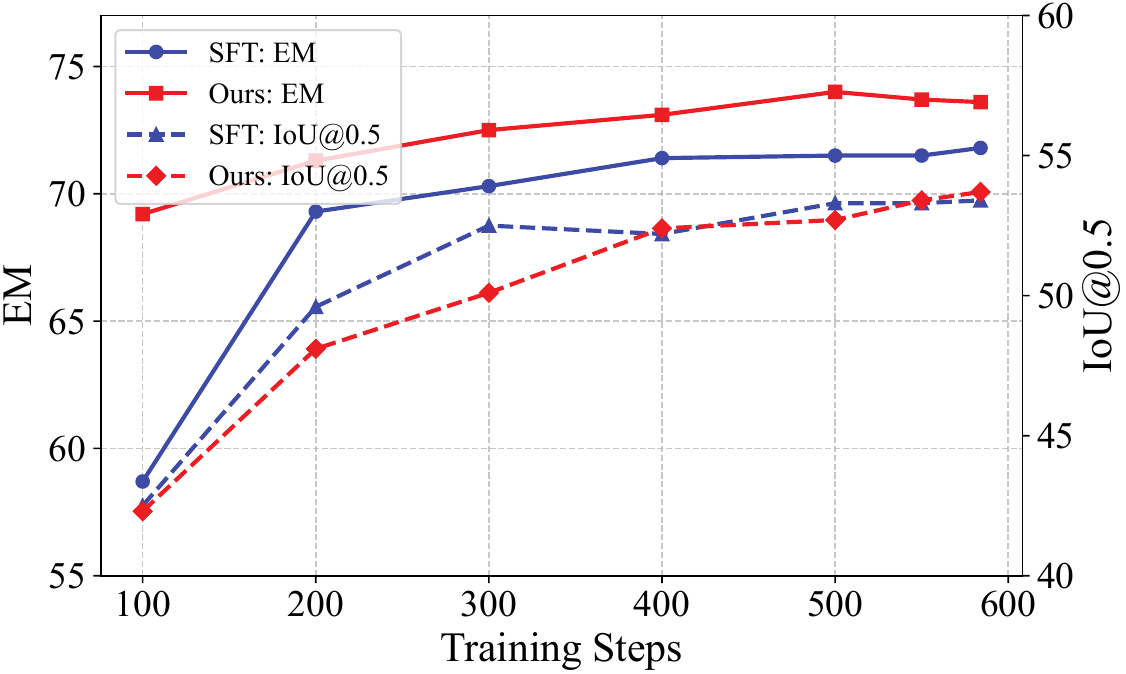}
        \caption{LAT vs. SFT. (Wiki-VISA Single)}
        \label{fig:sftvsrl}
    \end{subfigure}
    \hfill
    \begin{subfigure}[b]{0.32\textwidth}
        \centering
        \includegraphics[width=\textwidth]{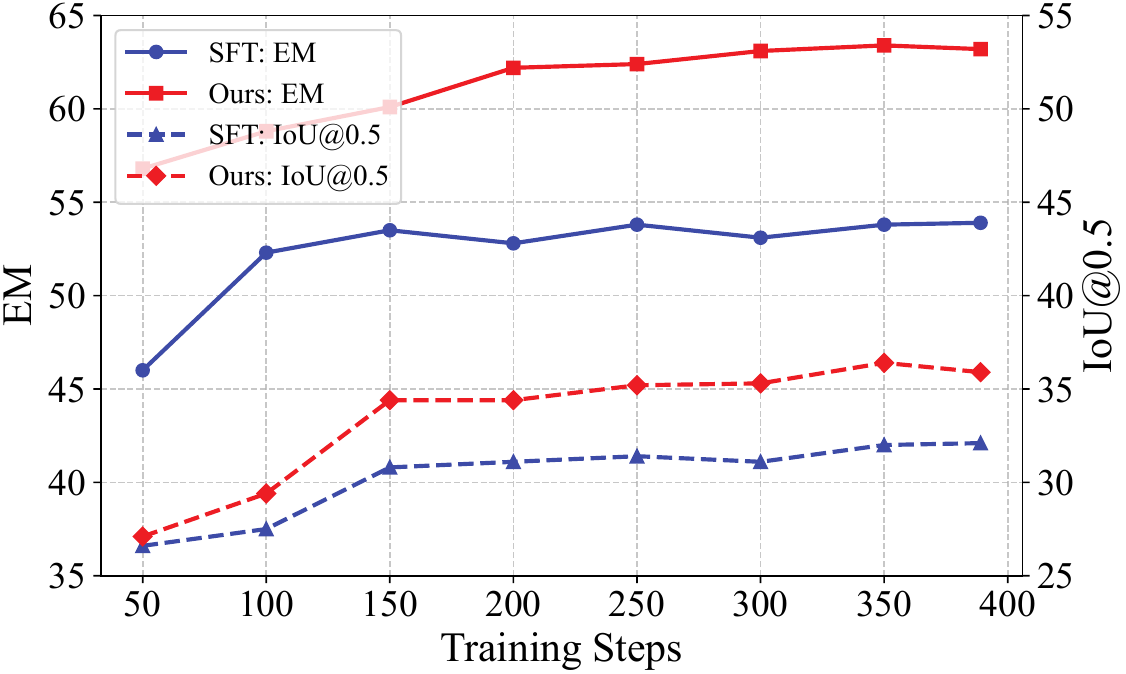}
        \caption{LAT vs. SFT. (Wiki-VISA Multi)}
        \label{fig:sftvsrl2}
    \end{subfigure}
    \hfill
    \begin{subfigure}[b]{0.32\textwidth}
        \centering
        \includegraphics[width=\textwidth]{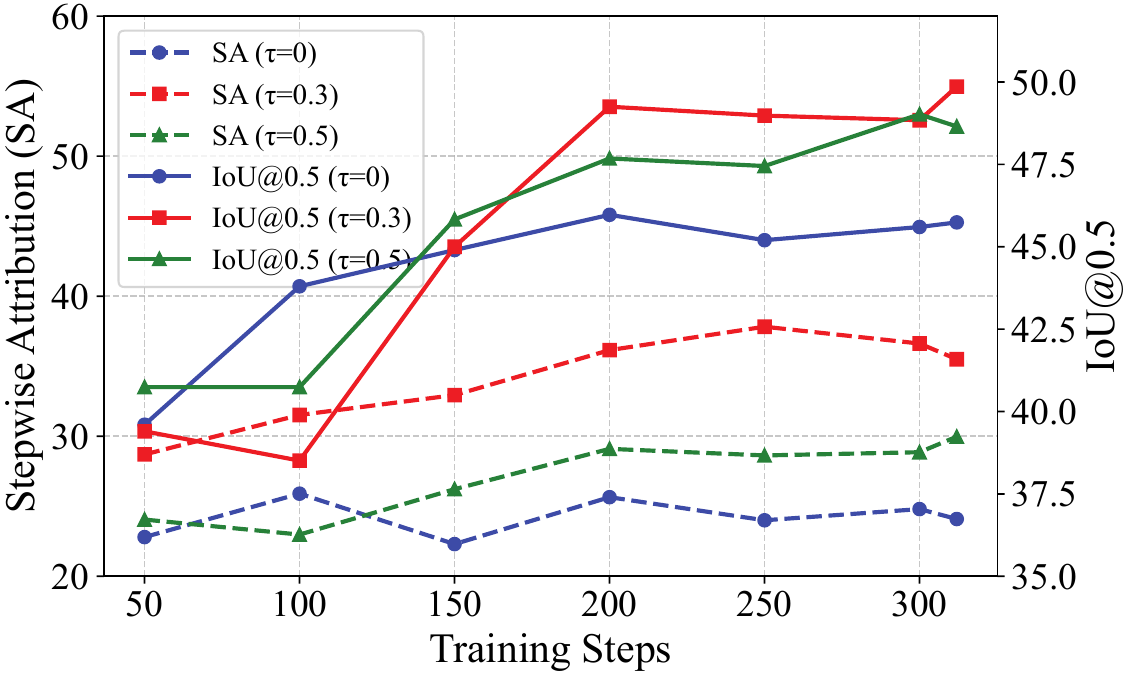}
        \caption{Performance variation with different $\tau$.}
        \label{fig:tau_ablation}
    \end{subfigure}
    \caption{Comparison of LAT and SFT performance across different settings and ablation study on threshold $\tau$.}
    \label{fig:combined_results}
\end{figure*}

\subsubsection{CoE reasoning performance with limited supervision.}
Unlike VISA, which relies on large-scale (100k) supervised data and directly links the final answer to supporting evidence without reasoning, LAT is trained on only 5\% of raw QA pairs during the RL stage. In low-resource settings, LAT achieves comparable performance to VISA-7B in answer accuracy and attribution precision, while maintaining traceable CoE reasoning. Notably, on high-resolution Wiki-VISA images, LAT outperforms VISA-7B, demonstrating robustness in visually complex scenarios under limited supervision.

To ensure fair comparison, we established an SFT baseline by training the model with VISA's supervision protocol on the same data subset and experimental setup used for our approach. Figure~\ref{fig:sftvsrl} demonstrates LAT's superior performance in both answer accuracy and evidence precision, with greater improvements observed in the multi-image setting (Figure~\ref{fig:sftvsrl2}). Meanwhile, Table~\ref{tab:general} highlights LAT's generalization across datasets, outperforming SFT by 1.7\% in EM and 6.2\% in IoU@0.5 in the ``Paper$\rightarrow$Wiki'' transfer setting, while preserving the vanilla model’s performance on Wiki- (EM: 67.7\%) and Paper-VISA (EM: 38.2\%). This demonstrates that LAT improves transfer effectiveness without sacrificing adaptability to diverse data types.

\begin{table}[t]
\centering
\small
\setlength{\tabcolsep}{1mm}
\begin{tabular}{cclll}
\toprule
\textbf{Train$\rightarrow$Eval} & \textbf{Method} & EM & IoU@0.5 \\
\midrule
\multirow{2}{*}{Paper$\rightarrow$Wiki (Single)} 
  & SFT & $66.0$ & $29.4$ \\
  & LAT-Ind. & \text{$67.7_{\uparrow1.7}$} & \text{$35.6_{\uparrow6.2}$} \\
\midrule
\multirow{2}{*}{Paper$\rightarrow$Wiki (Multi)} 
  & SFT & $48.7$ & $10.3$ \\
  & LAT-Ind. & \text{$57.3_{\uparrow8.6}$} & \text{$21.4_{\uparrow11.1}$} \\
\bottomrule
\end{tabular}
\caption{LAT demonstrates robust generalization with cross-domain transfer between Paper-VISA and Wiki-VISA.}
\label{tab:general}
\end{table}

\subsection{Ablation Study}

\begin{table*}[t]
\centering
\small
\setlength{\tabcolsep}{1mm}
\begin{tabular}{l|ccc|ccc|ccc|ccc}
\toprule
\textbf{Models} & \multicolumn{3}{c|}{\textbf{Wiki-VISA (Single)}} & \multicolumn{3}{c|}{\textbf{Paper-VISA (Single)}} & \multicolumn{3}{c|}{\textbf{Wiki-VISA (Multi)}} & \multicolumn{3}{c}{\textbf{Paper-VISA (Multi)}}\\
\cmidrule(lr){2-4} \cmidrule(lr){5-7} \cmidrule(lr){8-10} \cmidrule(lr){11-13}
& EM & IoU@0.5 & SA & EM & IoU@0.5 & SA & EM & IoU@0.5 & SA & EM & IoU@0.5 & SA \\
\midrule
Vanilla model (Qwen2.5-VL-7B) & 60.4 & 2.20 & 13.9 & 36.9 & 3.80 & 12.4 & 54.9 & 11.0 & 12.5 & 39.6 & 16.2 & 29.5\\
\midrule
$\mathcal{M}_\text{dist}$ (Stage I) & 67.1 & 25.6 & 32.8  & 44.7 & 27.0 & 13.0 & 59.4 & 27.8 & 44.0 & 46.5 & 36.3 & 28.2 \\
\textbf{LAT-Ind.} (Stage II) & \textbf{73.6} & \textbf{53.7} & \textbf{64.6} & \textbf{45.4} & \textbf{49.9} & 35.5 & \textbf{64.5} & \textbf{38.0} & 71.8 & \textbf{51.2} & \textbf{46.9} & 46.3 \\
w/o. $R_{\mathrm{step}}$ & 73.1 & 49.8 & 43.8 & 45.0 & 45.7 & 24.1 & 61.8 & 35.0 & 48.0 & 49.0 & 44.8 & 40.5\\
w/o. $R_\mathrm{acc}, R_\mathrm{ground}$ & 72.7 & 30.4 & 59.8 & 44.9 & 28.7 & 44.2 & 59.0 & 31.8 & 67.1 & 50.7 & 41.3 & 54.7\\
w/o. $R_\mathrm{acc}, R_\mathrm{ground}, (R_{\mathrm{acc}} \ge \epsilon)$ & 67.0 & 28.9 & 56.0 & 45.0 & 22.6 & \textbf{60.7} & 31.7 & 22.5 & \textbf{81.7} & 37.6 & 30.3 & \textbf{63.2} \\
\bottomrule
\end{tabular}
\caption{Ablation study results on LAT. ``w/o.'' denotes the results of models trained without the corresponding reward function.}
\label{tab:Ablation}
\end{table*}

\subsubsection{Effectiveness of reward components.} We conducted ablation studies on both datasets to examine the contributions of individual components in our reward formulation (Table~\ref{tab:Ablation}). We first analyzed the impact of distillation based on annotated CoE reasoning trajectories. The model $\mathcal{M}_\text{dist}$, obtained through fine-tuning during the cold start stage, shows an average improvement of \textbf{6.48\%} in EM and \textbf{20.9\%} in IoU@0.5 compared to the vanilla model. This indicates that distillation improves adherence to CoE reasoning formats. 

Next, we assess the impact of stepwise attribution by ablating the process reward $R_{\mathrm{step}}$. The model fails to align intermediate reasoning steps with visual evidence, resulting in a 15.5\% reduction in SA. Since the evidence grounding of the answer is inherently linked to the quality of intermediate visual attribution ($B_{ans} \in \mathcal{B}$), we also observed a decline in IoU@0.5, confirming the necessity of step-level attribution. Meanwhile, without the overlap constraint \( I \le \delta \) in $R_{\mathrm{step}}$, the model tends to reuse large and redundant regions across reasoning steps. This behavior undermines the coarse-to-fine grounding strategy and reduces attribution fidelity. 

\subsubsection{Aligning reasoning and result objectives.} We further analyzed the role of jointly optimizing process- and result-level rewards, where we removed the supervision from the final answer (w/o. $R_\mathrm{acc}, R_\mathrm{ground}$) and used it solely as a filter for valid reasoning paths. The model maintains moderate EM performance, indicating that consistency between the reasoning process and the answer inherently provides useful training signals. However, when supervision $R_{\mathrm{acc}} \ge \epsilon$ is removed from $R_{\mathrm{step}}$, the training objective is reduced to aligning only the reasoning format, resulting in significant degradation in both answer accuracy and evidence grounding precision. This suggests that $R_{\mathrm{step}}$ and result-level rewards work synergistically, where process rewards guide reasoning coherence while answer supervision ensures factual accuracy and proper grounding to source content. 

\subsection{Further Analysis}
\label{Analysis}

\subsubsection{Sensitivity analysis of attribution threshold $\tau$.}
To evaluate the alignment quality between visual evidence and textual references in CoE reasoning, we introduced a similarity threshold $\tau$ in the stepwise attribution reward function to distinguish positive from negative samples. Based on the synthetic CoE dataset $\mathcal{D}_{\text{final}}$, we computed semantic similarity scores across different answer types. As shown in Figure~\ref{fig:tau} in Appendix~\ref{tau analysis}, we reported the range of similarity values for each category, excluding pairs with zero scores. We set the default threshold to 0.3 based on the distribution analysis.

Given the parameter sensitivity, we further conducted experiments on both datasets to evaluate the robustness of $\tau$ across different document types. Specifically, we compared a high-threshold setting ($\tau{=}0.5$), a no-step variant (equivalent to $\tau{=}0$ or $1$, where all steps are uniformly rewarded), and the default setting. As shown in Figure~\ref{fig:tau_ablation} and Figure~\ref{fig:tau_ablation2} (Appendix~\ref{tau analysis}), $\tau{=}0.3$ achieves a better balance, yielding consistent improvements in both IoU@0.5 and SA throughout training. In contrast, the high threshold fails to sustain performance gains, as such strict criteria make it difficult to sample sufficient positive instances, while the no-step variant underperforms due to the lack of fine-grained attribution guidance. These results suggest that attribution supervision at $\tau{=}0.3$ offers relatively effective guidance for stepwise grounding. Additional analyses are reported in Appendix~\ref{tau analysis}.

\subsubsection{Traceable Reasoning with the CoE Paradigm.}
The CoE paradigm achieves stepwise visual attribution, generating a traceable reasoning process toward the final answer. Through the LAT framework, we leverage stepwise rewards to achieve an average SA of 57.1\%. Meanwhile, by penalizing repetitive reasoning processes in Equation~\ref{eq:IoU_formula}, we encourage the model to generate diverse and fine-grained reasoning. As illustrated in Figure~\ref {fig:case1}--\ref{fig:case6} in Appendix~\ref{sec:cs}, LAT accurately identifies the answer regions and generates faithful reasoning paths that closely align with the visual evidence.

\section{Conclusion}
\label{sec:bibtex}
In this paper, we introduce Chain-of-Evidence (CoE), a reasoning paradigm that unifies CoT with stepwise visual evidence attribution. To achieve CoE, we propose Look As You Think (LAT), a reinforcement learning framework that aligns the intermediate process to mitigate ungrounded reasoning for the visual evidence attribution task in VD-RAG. By incorporating stepwise rewards under the GRPO algorithm, LAT facilitates verification at each reasoning step. Experiments on Paper- and Wiki-VISA show that LAT outperforms baselines. We hope this work inspires further research on enhancing the verifiability of VD-RAG systems.

\section*{Acknowledgements}
This work was supported in part by the grants from National Science and Technology Major Project (No. 2023ZD0121104), and National Natural Science Foundation of China (No.62222213, 62072423).

\bibliography{aaai2026}

\clearpage
\newpage
\appendix
\twocolumn[
\begin{center}
\Large \textbf{Look As You Think: Unifying Reasoning and Visual Evidence Attribution for Verifiable Document RAG via Reinforcement Learning}\\
\vspace{+1em}
\large Supplementary Material
\end{center}
\vspace{+1em}
]

Visual evidence attribution in visual document retrieval-augmented generation (VD-RAG) requires models to both generate accurate answers and identify supporting evidence within external visual documents. 

We propose the LAT framework, which enables Chain-of-Evidence (CoE) reasoning through a two-stage training paradigm. In Stage I, the model is fine-tuned on a set of human-verified CoE annotations to acquire effective reasoning patterns. Next, it undergoes reinforcement learning (RL) with tailored reward functions that guide stepwise attribution. This design enhances both reasoning capability and attribution quality, thereby achieving robust stepwise attribution and consistent performance across single- and multi-image scenarios without relying on process-level ground-truth annotations.

\section{Experiment Details}
\label{adx:Details}

Since CoE reasoning requires flexible grounding capabilities and adaptability in low-resource conditions, we conducted preliminary experiments to identify a suitable backbone model. Specifically, we evaluated Qwen2.5-VL and InternVL2.5 by prompting them to answer questions with direct evidence attribution. As shown in Table~\ref{tab:direct_answer_result2}, Qwen2.5-VL achieved a 2.1\% improvement in EM under zero-shot prompting, whereas InternVL2.5 exhibited performance degradation. The weaker performance of InternVL2.5 may stem from its reliance on special tokens (e.g., \texttt{<box>}, \texttt{<ref>}) for grounding, which constrains the flexible grounding necessary for CoE reasoning in zero-shot settings and impedes effective convergence when training under low-resource conditions.

In contrast, Qwen2.5-VL represents bounding boxes in JSON format, allowing more flexible and direct visual evidence attribution without the constraints of special tokens. This design aligns well with the requirements of CoE reasoning. Building on this foundation, we introduce the two-stage training framework to enhance evidence grounding and reasoning consistency.

\paragraph{Stage I Cold Start:} We sampled instances from each dataset (Paper-VISA, Wiki-VISA, FineWeb-VISA) and filtered them using the recall metric defined in Section~\ref{sec:Proposed Approach} to retain correct samples. Manual correction was then performed to address bounding box drift by adjusting their positions and sizes to ensure alignment with the correct content regions. The final dataset splits are summarized in Table~\ref{tab:visa-dataset-stats} (cf. Table~\ref{tab:visa-dataset-full} for data usage comparison). We employed LoRA for parameter-efficient fine-tuning and maintained these configurations throughout training. 

\begin{table}[t]
\centering
\small
\begin{tabular}{lcc}
\toprule
Dataset        & \# Train-VISA & \# Test \\ 
\midrule
Wiki-VISA      & 87k      & 3,000   \\
Paper-VISA     & 100k     & 2,160   \\
Fineweb-VISA   & 60k      & -       \\
\bottomrule
\end{tabular}
\caption{Statistics for the entire VISA dataset, where the construction quantity for multi-image settings is consistent with that for single-image settings.}
\label{tab:visa-dataset-full}
\end{table}

\begin{table}[t]
  \centering
  \small
  \setlength{\tabcolsep}{1mm}
  \begin{tabular}{lccc}
    \toprule
    \multirow{2}{*}{Dataset} & \multicolumn{2}{c}{\# Train-LAT} & \multirow{2}{*}{\# Test} \\
    \cmidrule(lr){2-3}
     & Cold-Start Stage & RL Stage & \\      
    \midrule
    \multicolumn{4}{c}{\textbf{Single-Image / Multi-Image}} \\[0.3ex]
    Wiki-VISA  & 264 / 463 & 4,676 / 4,676 & 3,000 \\
    Paper-VISA & 271 / 355 & 5,000 / 3,562   & 2,160 \\
    Fineweb-VISA & 108 / 124 & 2,000 / 2,000   & - \\
    \bottomrule
  \end{tabular}
  \caption{Training subsets sampled from the Paper- and Wiki-VISA datasets under single- and multi-image settings.}
  \label{tab:visa-dataset-stats}
\end{table}

\paragraph{Stage II RL Training:} We employed GRPO as the policy optimization framework, with the reward function defined in Section~\ref{reward_design}. To ensure training stability, the learning rate was set to $5\mathrm{e}{-5}$ for both Paper- and Wiki-VISA, and training was conducted for one epoch. We further configured 8 and 6 rollouts for Paper- and Wiki-VISA, respectively, to enhance sampling diversity. The maximum prompt and completion lengths were limited to 16,384 and 600 tokens.

For the stepwise attribution reward $R_{\mathrm{step}}$, we required $R_{\mathrm{acc}} \ge \epsilon$, which zeroes the step rewards for incorrect answers and thereby reduces the advantage of erroneous samples. As specified in Equation~\ref{eq:acc}, if the predicted answer $a$ exactly matches the ground truth $a_{gt}$ (EM = 1), the recall is 1 and the reward is set to 1. For non-exact matches, recall was computed as the proportion of overlapping tokens between $a$ and $a_{gt}$, normalized by the ground truth length (Equation~\ref{eq:recall}). Manual evaluation of a representative subset indicates that answers require at least 80\% token overlap to preserve semantic completeness ($\gamma=0.8$). This corresponds to the accuracy reward $R_{acc}$ of approximately 0.4 ($\frac{\mathbb{I}\left(\mathrm{EM}(a, a_{gt}) = 1\right) + \text{Recall}(a, a_{gt})}{2} = \frac{0 + 0.8}{2}$). Accordingly, we set $\epsilon=0.4$ as the threshold, ensuring that faithful reasoning towards the answer is rewarded.  

To promote attribution diversity across reasoning steps, we applied IoU-based constraints with a threshold $\gamma=0.5$. A predicted box was regarded as valid evidence if its IoU with the corresponding ground truth exceeded 0.5, indicating sufficient relevance. Conversely, pairwise IoU among predicted boxes was required to remain below 0.5, ensuring that distinct reasoning steps relied on independent, non-overlapping evidence regions.

\paragraph{Implementation Details:} All experiments are conducted using Qwen2.5-VL-7B-Instruct on 4×A800 (80 GB) GPUs, requiring approximately 20 GPU hours for single-domain training and 48 GPU hours for multi-domain settings. Training leverages mixed precision with DeepSpeed~\cite{rasley2020deepspeed, rajbhandari2021zero} and CPU offloading for memory efficiency. To mitigate computational overhead, we employed Flash-attention2~\cite{dao2023flashattention2fasterattentionbetter} and gradient checkpointing techniques. Detailed hyperparameters for different training stages are presented in Table~\ref{tab:hyperparameters}. The pseudocode for LAT is provided in Algorithm~\ref{alg:lat_training}. For multi-image scenarios, initialize the model from the single-image RL-trained parameters, then further apply SFT on multi-image CoE data in $\mathcal{D}_{\text{final}}$ following the above procedures.

\begin{table}[t]
    \centering
    \small
    \begin{tabular}{@{}lll@{}}
        \toprule
        \textbf{Stage} & \textbf{Parameter} & \textbf{Value} \\
        \midrule
        \multirow{7}{*}{Cold Start} 
        & $\gamma$ & 0.8 \\
        & learning rate & $1\mathrm{e}{-4}$ \\
        & batch Size & 8 \\
        & epoch(s) & 2 \\
        & lora rank (\(r\)) & 64 \\
        & lora scaling (\(\alpha\)) & 64 \\
        & lora dropout & 0.05 \\
        \midrule
        \multirow{10}{*}{RL Training} 
        & learning rate & $5\mathrm{e}{-5}$ \\
        & batch size & 16 / 8 \\
        & epoch(s) & 1 \\
        & rollout & 8 / 6 \\
        & $\tau$ & 0.3 \\
        & $\delta$ & 0.5 \\
        & $\epsilon$ & 0.4 \\
        & lora rank (\(r\)) & 64 \\
        & lora scaling (\(\alpha\)) & 64 \\
        & lora dropout & 0.05 \\
        \bottomrule
    \end{tabular}
    \caption{Hyperparameter Settings for Training. Values marked with A / B correspond to different configurations used for Paper- and Wiki-VISA, respectively.}
    \label{tab:hyperparameters}
\end{table}

\begin{algorithm}[tb]
\small
\caption{LAT Framework}
\label{alg:lat_training}
\textbf{Input}: Sampled query set $\mathcal{Q}_{\text{cold\_start}}$ and $\mathcal{Q}_{\text{RL}}$ \\
\textbf{Parameter}: Learning rates $\eta_1$, $\eta_2$; Reward thresholds $\tau, \delta, \epsilon, \gamma$ \\
\textbf{Output}: LAT model $\pi_\theta$

\begin{algorithmic}[1]
\STATE \textbf{Stage I: Cold-start Supervised Fine-tuning}
\FOR{each query $q \in \mathcal{Q}_{\text{cold\_start}}$}
    \STATE Generate CoE reasoning traces using Gemini2.5 pro
    \STATE Filter outputs by the recall metric
    \IF{$\text{Recall}(a, a_{gt}) \geq \gamma$}
    \STATE $\mathcal{D}_{\text{final}} \leftarrow \mathcal{D}_{\text{final}} \cup \{q\}$
    \ENDIF
    
\ENDFOR
\STATE Manually verify bounding boxes and format
\STATE Train model $\pi_\theta$ via SFT on verified data $\mathcal{D}_{\text{final}}$ with learning rate $\eta_1$

\STATE \textbf{Stage II: Reinforcement Learning with GRPO}
\FOR{each training step}
    \STATE Sample a batch of queries $q \sim \mathcal{Q}_{\text{RL}}$
    \STATE Model $\pi_\theta$ generates CoE reasoning steps $\{r_1, ..., r_T\}$ and bounding boxes $\{B_1, ..., B_T\}$ for each query $q$
    \STATE Compute reward $R = R_{\text{acc}} + R_{\text{step}} + R_{\text{ground}} + R_{\text{format}}$
    \STATE Update policy $\pi_\theta$ using GRPO with $R$ and learning rate $\eta_2$
\ENDFOR
\STATE \textbf{return} final LAT model $\pi_\theta$
\end{algorithmic}
\end{algorithm}

For reproducibility, we fixed the random seed to 3407 during training and disabled sampling at evaluation by setting \texttt{do\_sample=False}. In contrast to VISA~\cite{ma2024visaretrievalaugmentedgeneration}, we maintained original image resolutions since our CoE reasoning framework requires comprehensive visual understanding across the entire image. 


\section{Dataset Construction}
\label{data col}
\subsection{VISA Dateset}

\noindent \textbf{1) Wiki-VISA}: Selenium renders Wikipedia pages for Natural Questions (NQ)~\cite{kwiatkowski2019natural} query-answer pairs, with HTML elements (containing answers) and their bounding boxes annotated.

\noindent \textbf{2) Paper-VISA}: Based on PubLayNet~\cite{zhong2019publaynet}. Vision-language models (VLMs) generate QA pairs from layout-annotated scientific documents, with answer-region bounding boxes extracted.

\noindent \textbf{3) FineWeb-VISA}: Sampled from FineWeb-edu~\cite{penedo2024fineweb}. VLMs generate queries/short answers for educational webpage passages longer than 50 tokens, with screenshots and answer-region bounding boxes.

\subsection{Multi-Image Data Construction}
To simulate real-world VD-RAG scenarios, VISA constructs a multi-image document experimental environment after obtaining the query-document-answer-bounding box triplets. Specifically, given a query $q$, VISA employs a retriever to retrieve top-$k$ candidate documents, then randomly samples $m{-}1$ hard negative candidates that do not contain the ground truth. These negative samples are combined with one source document containing the correct answer to serve as input for the multi-image scenarios. VISA deliberately avoids directly utilizing the top-$m$ retrieval results to prevent bias toward specific retrievers or candidate document positions, thereby ensuring methodological generalizability. 

\begin{figure*}[t] 
\centerline{
\includegraphics[width=\textwidth]{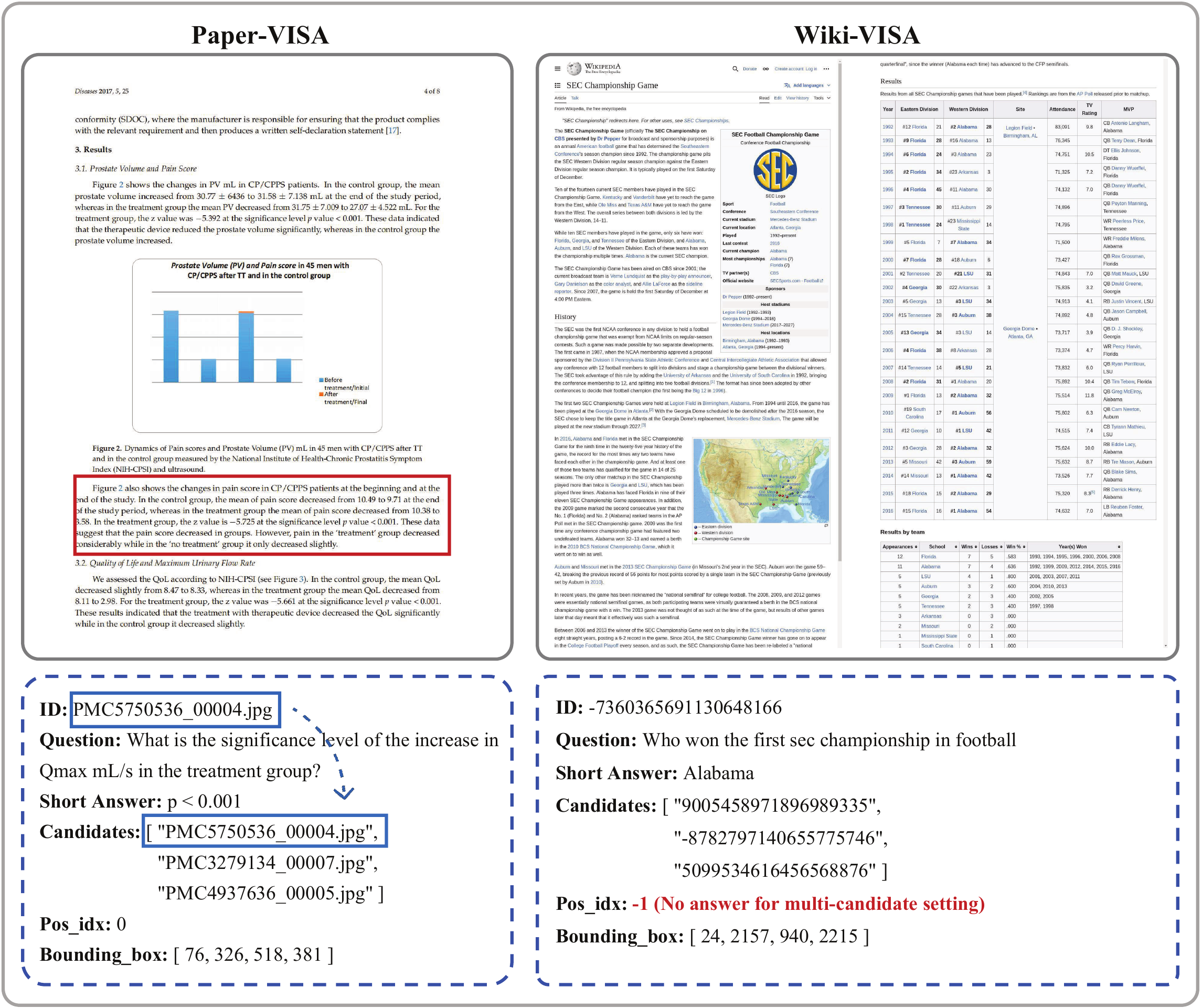} } 
\caption{ 
Data examples from Paper-VISA (left) and Wiki-VISA (right). Each image is assigned a unique identifier, with every dataset entry containing a reference image paired with a specific question, ground-truth answer, and answer source localized by a bounding box. In multi-image scenarios, a retriever selects two images and appends their IDs to the reference image, forming a candidate list. For example, the red bounding box (left) indicates the answer source, where \texttt{pos\_idx=0} signifies that the reference image occupies the first position in the candidate list. For entries lacking ground-truth answers (right), the reference image is substituted with an irrelevant image in the candidate list (\texttt{pos\_idx=-1}).
} 
\label{fig:visa} 
\end{figure*}

To evaluate the model's capability in handling no-answer scenarios, VISA randomly replaces the source document in the candidate set with a 20\% probability, simulating realistic cases where the retriever fails to return documents containing the correct answer. In the specific experimental setup, VISA leverages the Document Screenshot Embedding (DSE) model as the retriever. The parameters are set to $k=20$ and $m=3$. Example cases are shown in Figure~\ref{fig:visa}.

\begin{table*}[ht]
\centering
\small
\setlength{\tabcolsep}{1mm}
\begin{tabular}{l|c|cc|cc|cc|cc}
\toprule
\textbf{Models} & \textbf{Param.} & \multicolumn{2}{c|}{\textbf{Wiki-VISA (Single)}} & \multicolumn{2}{c|}{\textbf{Paper-VISA (Single)}} & \multicolumn{2}{c|}{\textbf{Wiki-VISA (Multi)}} & \multicolumn{2}{c}{\textbf{Paper-VISA (Multi)}} \\
\cmidrule(lr){3-4} \cmidrule(lr){5-6} \cmidrule(lr){7-8} \cmidrule(lr){9-10}
& \textbf{Size} & EM & IoU@0.5 & EM & IoU@0.5 & EM & IoU@0.5 & EM & IoU@0.5 \\
\midrule
\multicolumn{1}{c|}{} & \multicolumn{1}{c|}{} & \multicolumn{8}{c}{\textbf{Open-source Models, Direct Answer}} \\
InternVL2.5 & 8B & 45.9 & - & 42.4 & - & 37.7 & - & 32.1 & -\\
Qwen2.5-VL & 7B & 67.7 & - & 38.2 & - & 54.1 & - & 34.6 & - \\
\midrule
\multicolumn{1}{c|}{} & \multicolumn{1}{c|}{} & \multicolumn{8}{c}{\textbf{Open-source Models, Direct Answer \& Attribution (DA)}} \\
InternVL2.5 & 8B & 45.1 & 0 & 42.0 & 0.60 & 35.0 & 1.67 & 31.4 & 1.94\\
Qwen2.5-VL & 7B & 69.4 & 0.80 & 43.8 & 2.87 & 52.5 & 0.97 & 37.2 & 5.56 \\
\bottomrule
\end{tabular}
\caption{Performance comparison of various models for direct answer generation and attribution tasks. ``DA'' refers to the zero-shot prompting setting for direct answer \& attribution without training.}
\label{tab:direct_answer_result2}
\end{table*}

\subsection{Training Data Preprocess}
\begin{figure*}[t]
\centerline{
\includegraphics[width=\textwidth]{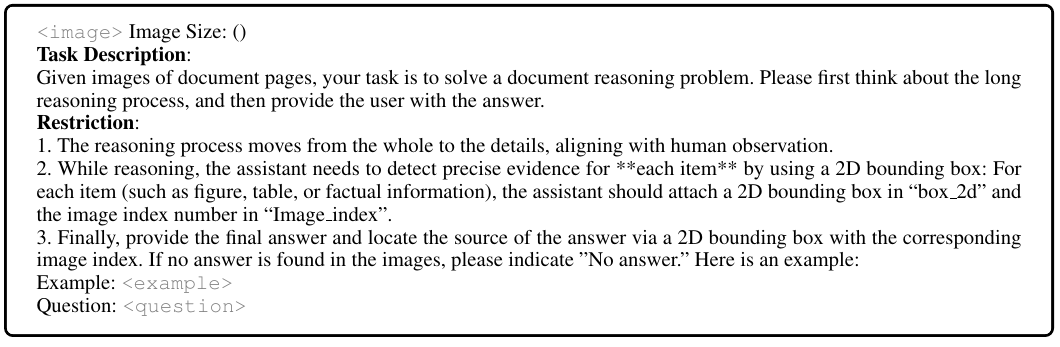} } 
\caption{Prompt template used for CoE Generation with Gemini2.5 pro.}
\label{fig:prompt_gemini-generation}
\end{figure*}

We employed Gemini2.5 Pro~\cite{comanici2025gemini25pushingfrontier} to generate CoE data. We used the prompt shown in Figure~\ref{fig:prompt_gemini-generation} to guide stepwise evidence attribution during generation.

\begin{figure*}[t] 
\centerline{
\includegraphics[width=\textwidth]{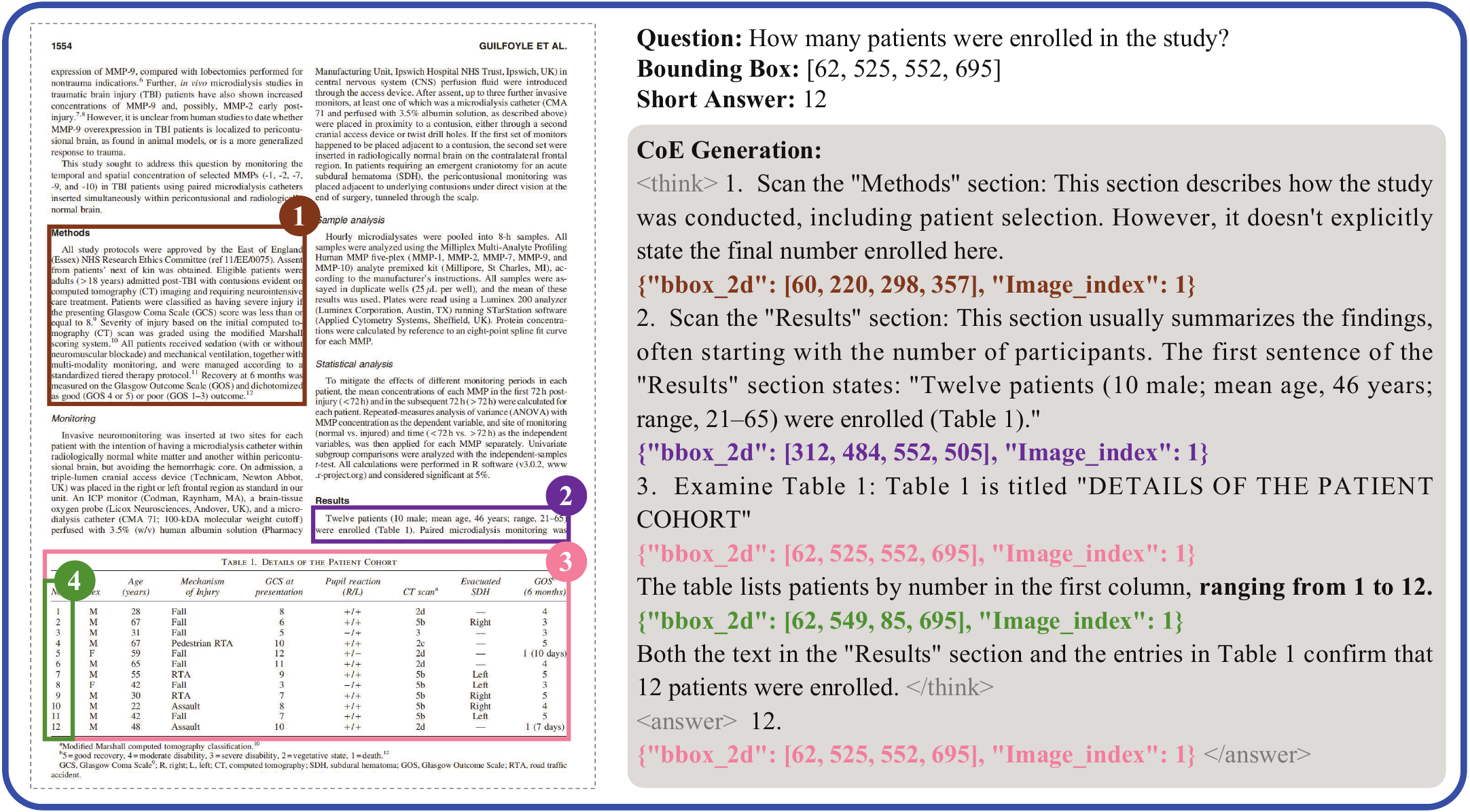} } 
\caption{ 
The Chain-of-Evidence data generated from the Paper-VISA during the cold-start phase.
} 
\label{fig:paper_coe} 
\end{figure*}

\begin{figure*}[t] 
\centerline{
\includegraphics[width=\textwidth]{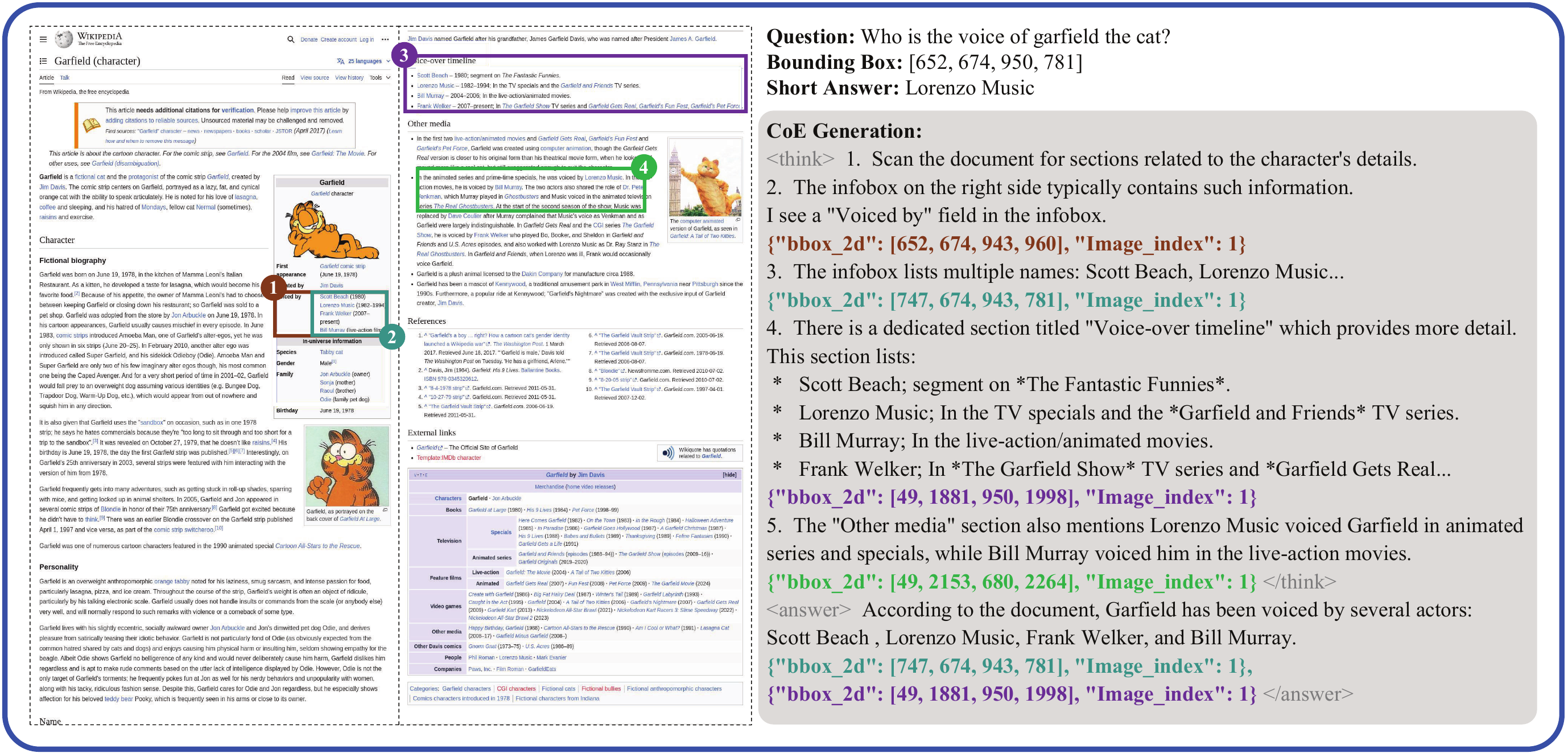} } 
\caption{ 
The Chain-of-Evidence data generated from the Wiki-VISA during the cold-start phase.
} 
\label{fig:wiki_coe} 
\end{figure*}

We subsequently applied a recall metric to measure response against reference answers for preliminary sample filtering. To ensure data quality, we manually reviewed and corrected instances of irrelevant attributions, spatial coordinate offsets, and format inconsistencies. The resulting dataset $\mathcal{D}_{\text{final}}$ is then employed for cold-start training. Examples are illustrated in Figure~\ref{fig:paper_coe} and~\ref{fig:wiki_coe}.

\section{Prompt}
\label{prompt}
We designed task-specific prompts adapted to reasoning strategies and model architectures. For models such as LLaVA-CoT~\cite{xu2025llavacotletvisionlanguage} and R1-OneVision~\cite{yang2025r1onevisionadvancinggeneralizedmultimodal}, which are trained to take questions directly as input, with no additional prompting. Their performance is evaluated by extracting content from the generated outputs, specifically from segments enclosed within the \texttt{<CONCLUSION>} and \texttt{<answer>} tags. Since both Gemini and InternVL generate bounding boxes in normalized coordinates. During evaluation, these normalized values are converted to absolute coordinates based on the size of the source images. For other models, we employed prompts that initiate responses with ``The answer is:'', which encourages concise answers without irrelevant explanations. 

\begin{figure*}[t]
\centerline{
\includegraphics[width=\textwidth]{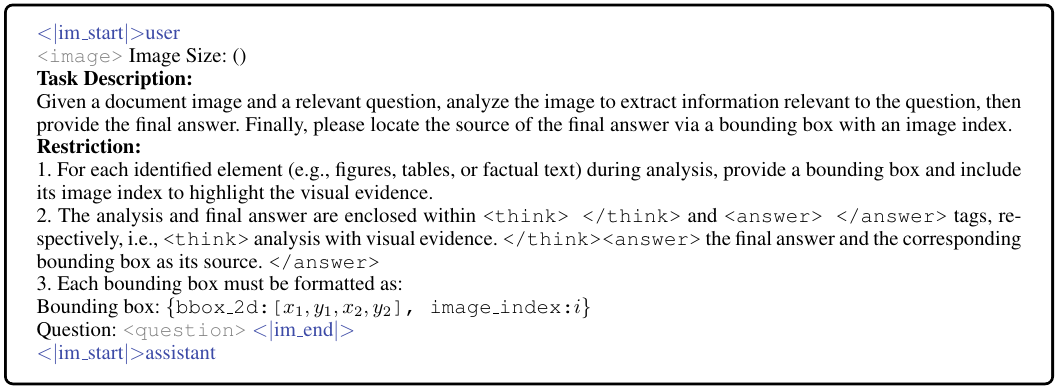} } 
\caption{Prompt template used for training and inference (Single-image) with Qwen2.5-VL. The structured prompt encourages attribution-aware reasoning.}
\label{fig:prompt_qwen-single}
\end{figure*}

Each bounding box must be paired with its source image index using the JSON format \texttt{\{"bbox\_2d": [x\_1, y\_1, x\_2, y\_2], "image\_index": i\}}. As illustrated in Figure~\ref{fig:prompt_qwen-single}, we employed a consistent prompt format across both the cold-start and RL training stages.

\begin{figure}[t] 
\centering
\includegraphics[width=0.35\textwidth]{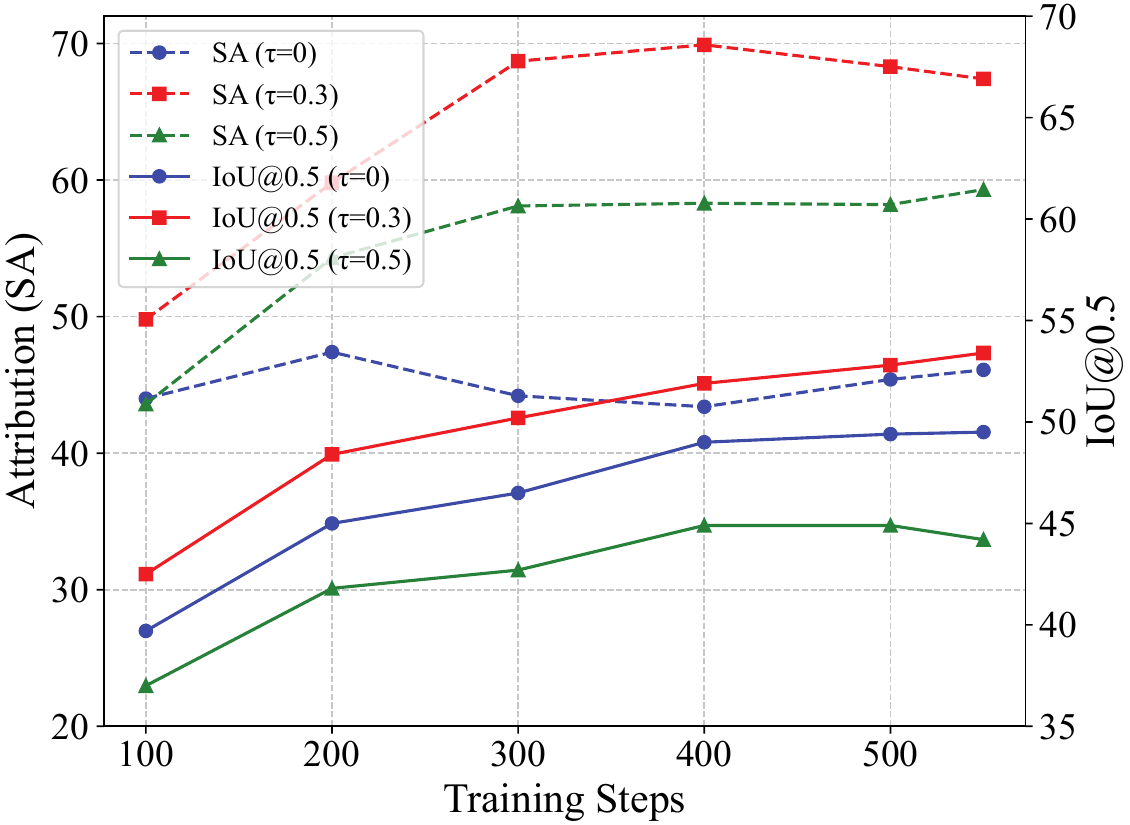}
\caption{Performance variation with different $\tau$ (Wiki).}
\label{fig:tau_ablation2}
\end{figure}

\begin{figure*}[t] 
\centerline{
\includegraphics[width=\textwidth]{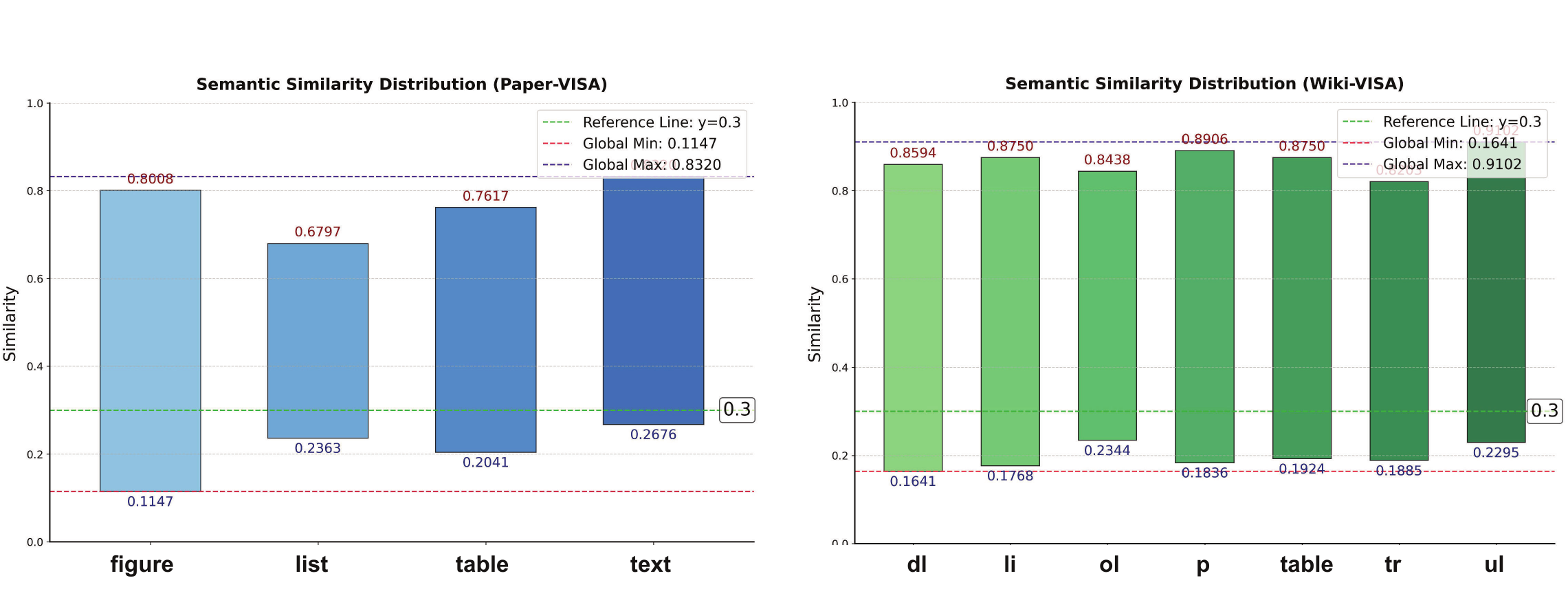} } 
\caption{ 
Semantic similarity distribution of synthetic CoE data. We computed the mean upper and lower bounds of semantic similarity between manually annotated evidence attributions and context-evidence pairs across different categories of synthetic CoE data in the cold-start stage. To ensure the effectiveness of the reward function, we selected a threshold $\tau=0.3$ in Equation~\ref{eq:threshold} to enforce well-aligned CoE reasoning steps during RL training.
} 
\label{fig:tau} 
\end{figure*}

\begin{figure*}[t]
\centerline{
\includegraphics[width=\textwidth]{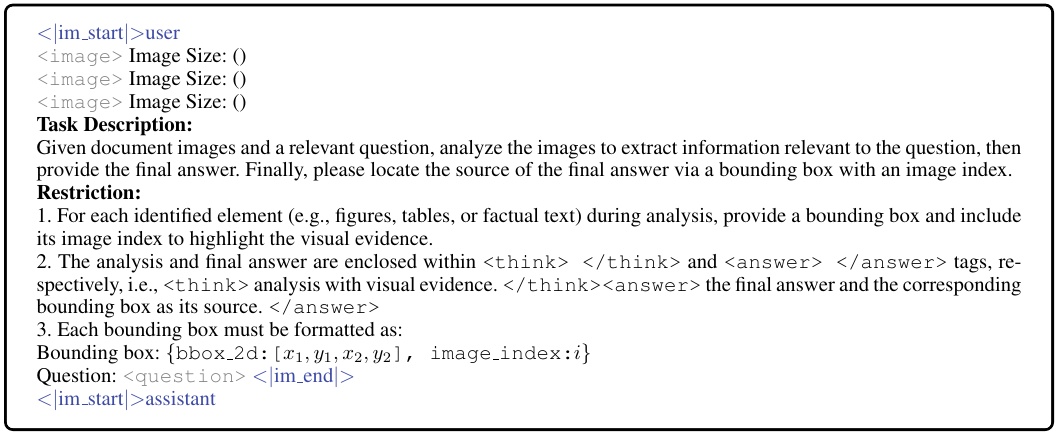} } 
\caption{Prompt template used for training and inference (Multi-image) with Qwen2.5-VL. The structured prompt encourages attribution-aware reasoning.}
\label{fig:prompt_qwen-multi}
\end{figure*}

\section{Baseline}
\label{baseline}
Our results demonstrate that LAT consistently outperforms open-source models of comparable or larger scales, including InternVL2.5-8B~\cite{chen2025expandingperformanceboundariesopensource}, Qwen2.5-VL-7/32B-Instruct~\cite{bai2025qwen25vltechnicalreport}, LLaVA-OneVision~\cite{li2024llavaonevisioneasyvisualtask}, mPLUG-DocOwl2~\cite{hu2024mplugdocowl2highresolutioncompressingocrfree}, LLaVA-CoT~\cite{xu2025llavacotletvisionlanguage}, and R1-OneVision~\cite{yang2025r1onevisionadvancinggeneralizedmultimodal}. We note that LLaVA-CoT is trained exclusively on single-image datasets without multi-image reasoning supervision and is therefore excluded from the multi-image evaluation.

LAT achieves superior performance compared to certain closed-source models, including Qwen-VL Max~\cite{bai2023qwenvlversatilevisionlanguagemodel} and Gemini 2.0 Flash~\cite{geminiteam2025geminifamilyhighlycapable}. In our experimental setting, VISA-7B is performed with the open-source implementation available on HuggingFace\footnote{https://huggingface.co/collections/MrLight/visa-rag-with-visual-source-attribution}. GCoT\cite{wu2025groundedchainofthoughtmultimodallarge} and VisCoT\cite{shao2024visual} rely on single-image datasets with costly step annotations (e.g., 438k) and are not trained in multi-image settings. The EM of VisCoT on Paper/Wiki-VISA is 10.5/8.8\%, illustrating the domain gap.

\section{Further Analysis}
\label{tau analysis}
\subsection{Sensitivity analysis of attribution threshold $\tau$.}

To ensure the alignment quality of visual evidence-text pairs in CoE reasoning, we set a threshold $\tau$ in the stepwise attribution recall reward function to distinguish positive from negative samples. Using the synthetic CoE data $\mathcal{D}_{\text{final}}$, we computed the values of semantic similarity on manually corrected annotations. As shown in Figure~\ref{fig:tau}, we recorded the maximum and minimum similarities for samples grouped by answer type in each dataset, excluding pairs with zero similarity. Based on this analysis, we selected 0.3 as the threshold, slightly above the minimum similarity observed in human-corrected annotations.

To examine the effect of different thresholds, we conducted ablation experiments on single-image scenarios from Paper-VISA (Figure~\ref{fig:tau_ablation}) and Wiki-VISA (Figure~\ref{fig:tau_ablation2}). These experiments reveal that the choice of $\tau$ has an impact on reasoning outcomes. For example, a relatively high threshold, such as 0.5, leads to notable declines in both accuracy and IoU@0.5, and the overall process traceability quality (SA) also fails to improve, because such strict thresholds make it difficult to sample sufficient positive instances. 

\begin{table}[t]
\centering
\small
\setlength{\tabcolsep}{1mm}
\begin{tabular}{cclll}
\toprule
\textbf{Train$\rightarrow$Eval} & \textbf{Method} & EM & IoU@0.5 \\
\midrule
\multirow{2}{*}{Wiki$\rightarrow$Paper (Single)} 
  & SFT & $36.9$ & $11.5$ \\
  & LAT-Ind. & \text{$43.6_{\uparrow6.7}$} & \text{$10.4_{\downarrow1.1}$} \\
\midrule
\multirow{2}{*}{Wiki$\rightarrow$Paper (Multi)} 
  & SFT & $34.3$ & $5.5$ \\
  & LAT-Ind. & \text{$44.3_{\uparrow10.0}$} & \text{$21.9_{\uparrow16.4}$} \\
\bottomrule
\end{tabular}
\caption{LAT demonstrates robust generalization with cross-domain transfer between Paper-VISA and Wiki-VISA.}
\label{tab:generalmulti}
\end{table}

\subsection{Generalization and Ablation}
The results in Table~\ref{tab:general} demonstrate LAT's cross-domain transferability, achieving superior performance over SFT with improvements of 1.7\% in EM and 6.2\% in IoU@0.5 when transferring from Paper- to Wiki-VISA datasets. In the reverse transfer (Wiki$\rightarrow$Paper), a slight drop in IoU@0.5 is observed. This can be attributed to the training bias on Wiki-VISA, which contains high-resolution images with relatively dispersed layouts, whereas Paper-VISA features compact medical documents. Such dense layouts often lead to incomplete evidence localization. Nevertheless, the model maintains strong reasoning consistency and answer accuracy, preserving the vanilla model’s performance on both Wiki-VISA (EM: 67.7\%) and Paper-VISA (EM: 38.2\%). The generalization is even more apparent in multi-image scenarios (Table~\ref{tab:general} and~\ref{tab:generalmulti}). These results indicate that LAT improves transfer effectiveness while retaining adaptability across diverse document types. 

We also conducted further ablation studies. Without the cold-start stage, the performance achievable by RL is fundamentally capped (EM, 43.9\%; IoU, 24.1\%; SA, 33.4\% on Paper-VISA). Moreover, the learning curves indicate that RL training for a single epoch has not yet reached its performance ceiling. We therefore extended the training to 2 epochs on Paper-VISA, achieving improved results (EM, \text{$46.5_{\uparrow1.1}\%$}; IoU, \text{$50.1_{\uparrow0.2}\%$}; SA, \text{$38.1_{\uparrow2.6}\%$}).

\begin{table}[t]
\centering
\small
\setlength{\tabcolsep}{1mm}
\begin{tabular}{llccc}
\toprule
\textbf{Dateset} & \textbf{Method} & Total & Correct & Acc. \\
\midrule
\multirow{2}{*}{Paper} 
  & LAT-Ind. & 425 & 356 & 0.84\\
  & LAT-Full & 425 & 343 & 0.81 \\
\midrule
\multirow{2}{*}{Wiki} 
  & LAT-Ind. & 578 & 267 & 0.46\\
  & LAT-Full & 578 & 283 & 0.49 \\
\bottomrule
\end{tabular}
\caption{The accuracy of correctly detecting no-answer cases in multi-image settings.}
\label{tab:general3}
\end{table}

\section{Case Study}
\label{sec:cs}

\begin{figure*}[t] 
\centerline{
\includegraphics[width=\textwidth]{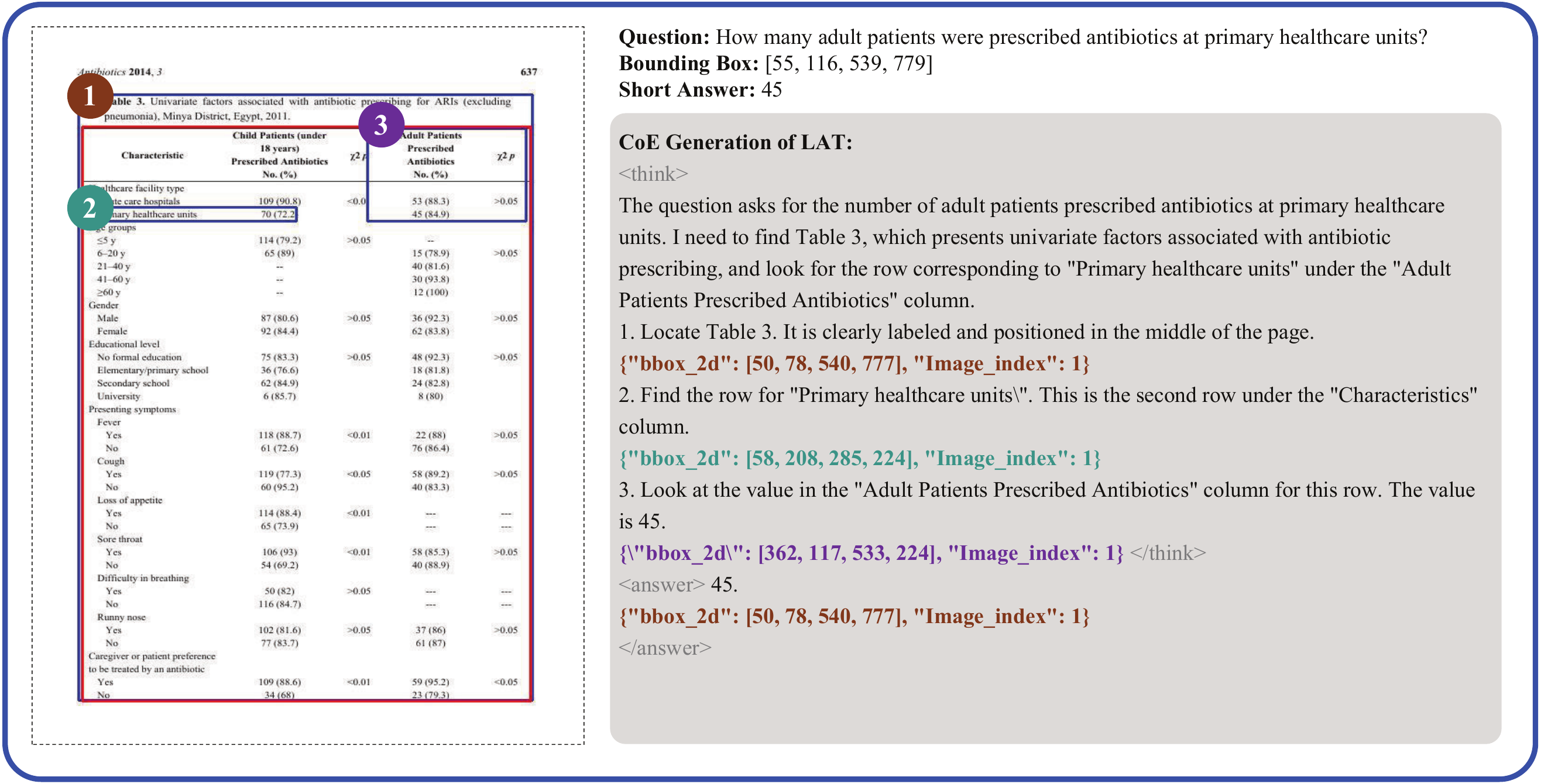} } 
\caption{ 
The Chain-of-Evidence reasoning process of LAT on the Paper-VISA. Blue boxes represent model predictions, while the red box indicates the ground truth bounding box from the dataset. (Case 1)
} 
\label{fig:case1} 
\end{figure*}

\begin{figure*}[t] 
\centerline{
\includegraphics[width=\textwidth]{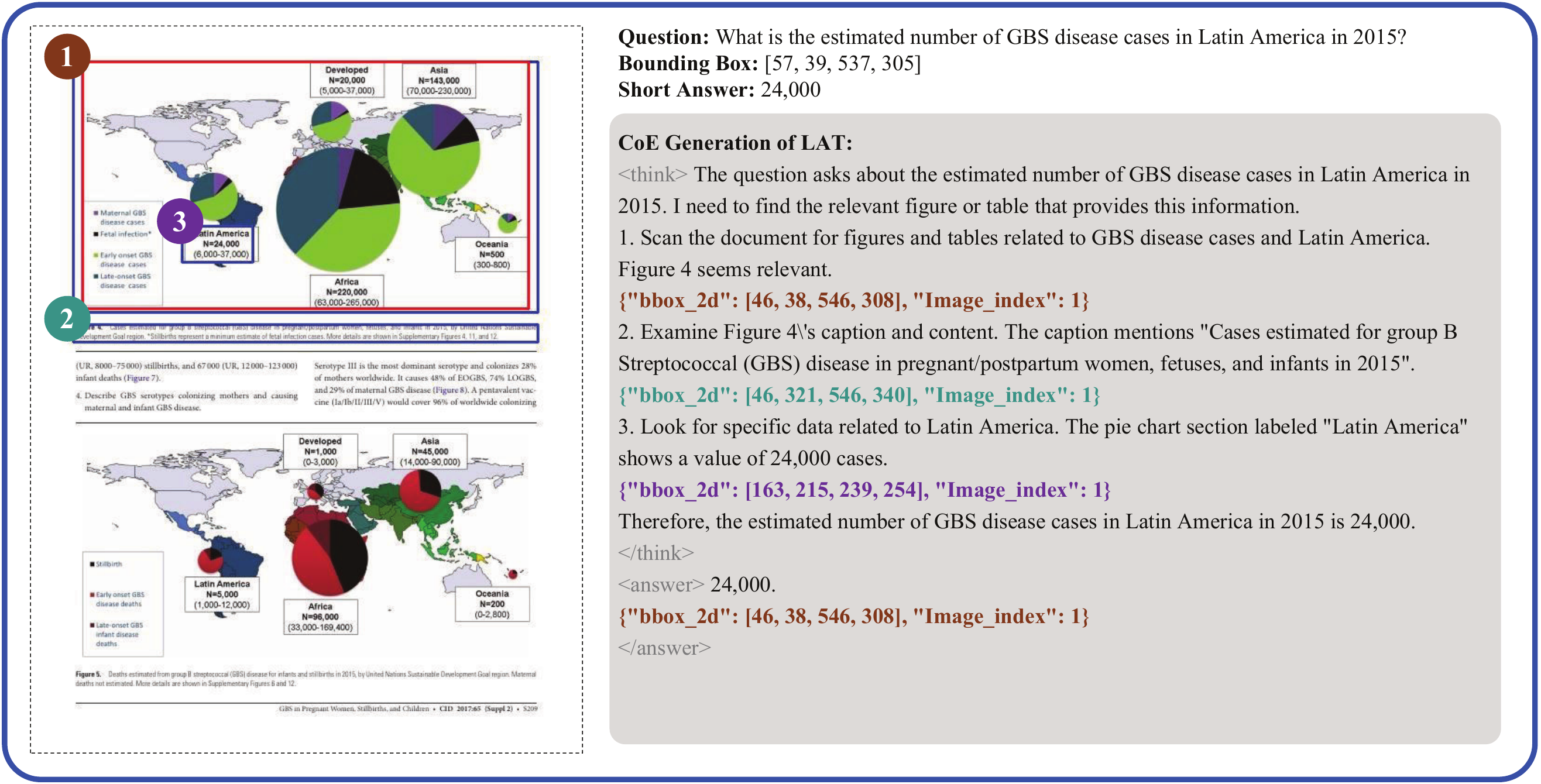} } 
\caption{ 
The Chain-of-Evidence reasoning process of LAT on the Paper-VISA. Blue boxes represent model predictions, while the red box indicates the ground truth bounding box from the dataset. (Case 2) 
} 
\label{fig:case2} 
\end{figure*}

\begin{figure*}[t] 
\centerline{
\includegraphics[width=\textwidth]{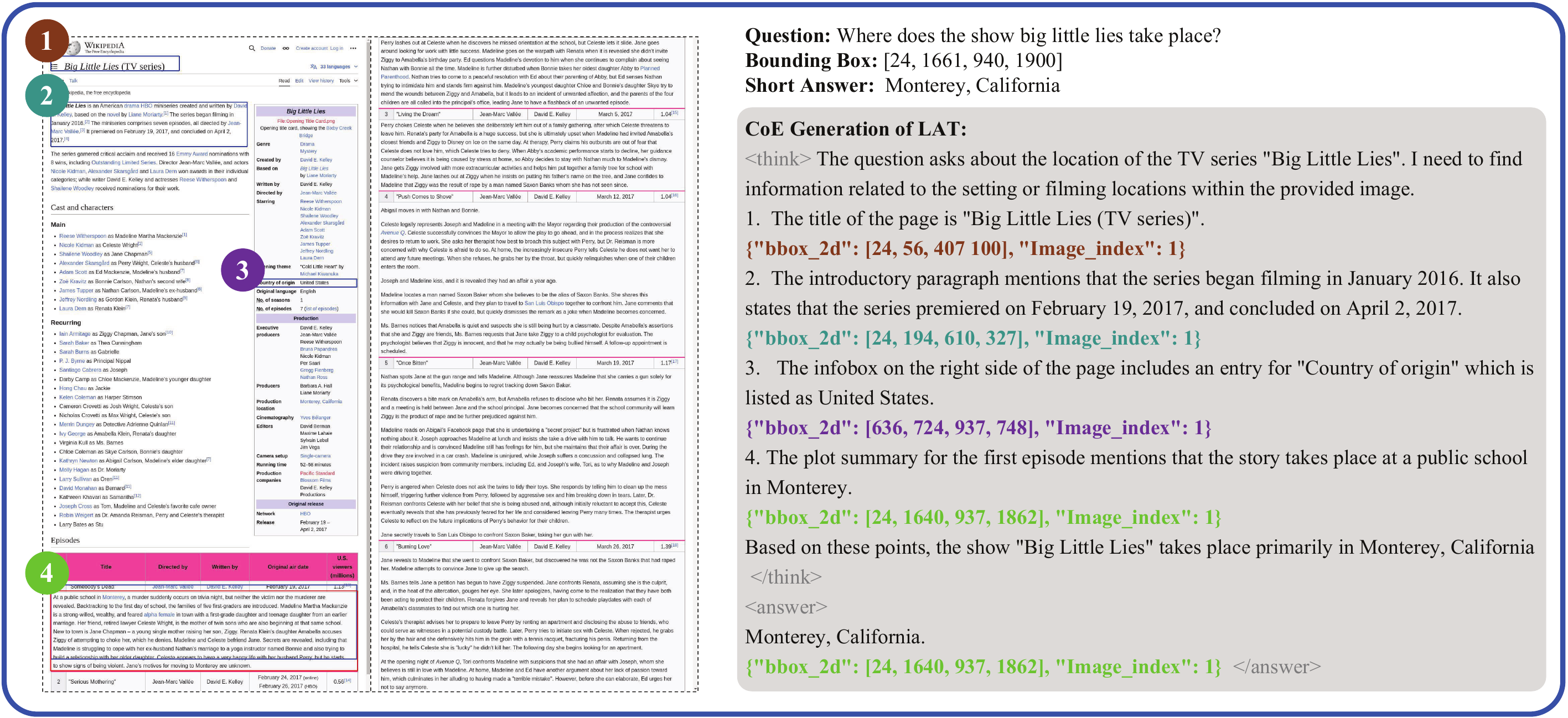} } 
\caption{ 
The Chain-of-Evidence reasoning process of LAT on the Wiki-VISA. Blue boxes represent model predictions, while the red box indicates the ground truth bounding box from the dataset. (Case 3)
} 
\label{fig:case3} 
\end{figure*}

\begin{figure*}[t] 
\centerline{
\includegraphics[width=\textwidth]{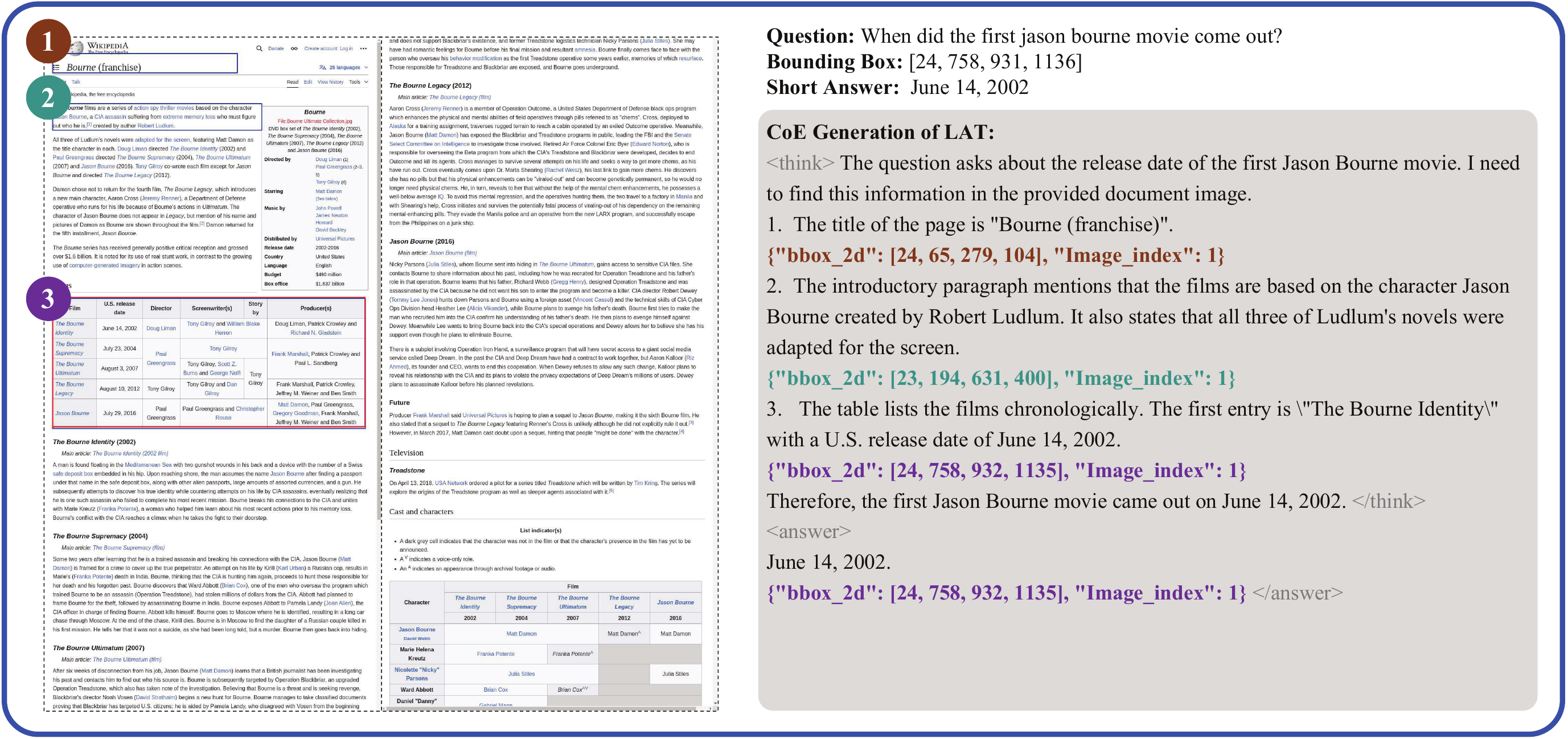} } 
\caption{ 
The Chain-of-Evidence reasoning process of LAT on the Wiki-VISA. Blue boxes represent model predictions, while the red box indicates the ground truth bounding box from the dataset. (Case 4)
} 
\label{fig:case4} 
\end{figure*}

\begin{figure*}[t] 
\centerline{
\includegraphics[width=\textwidth]{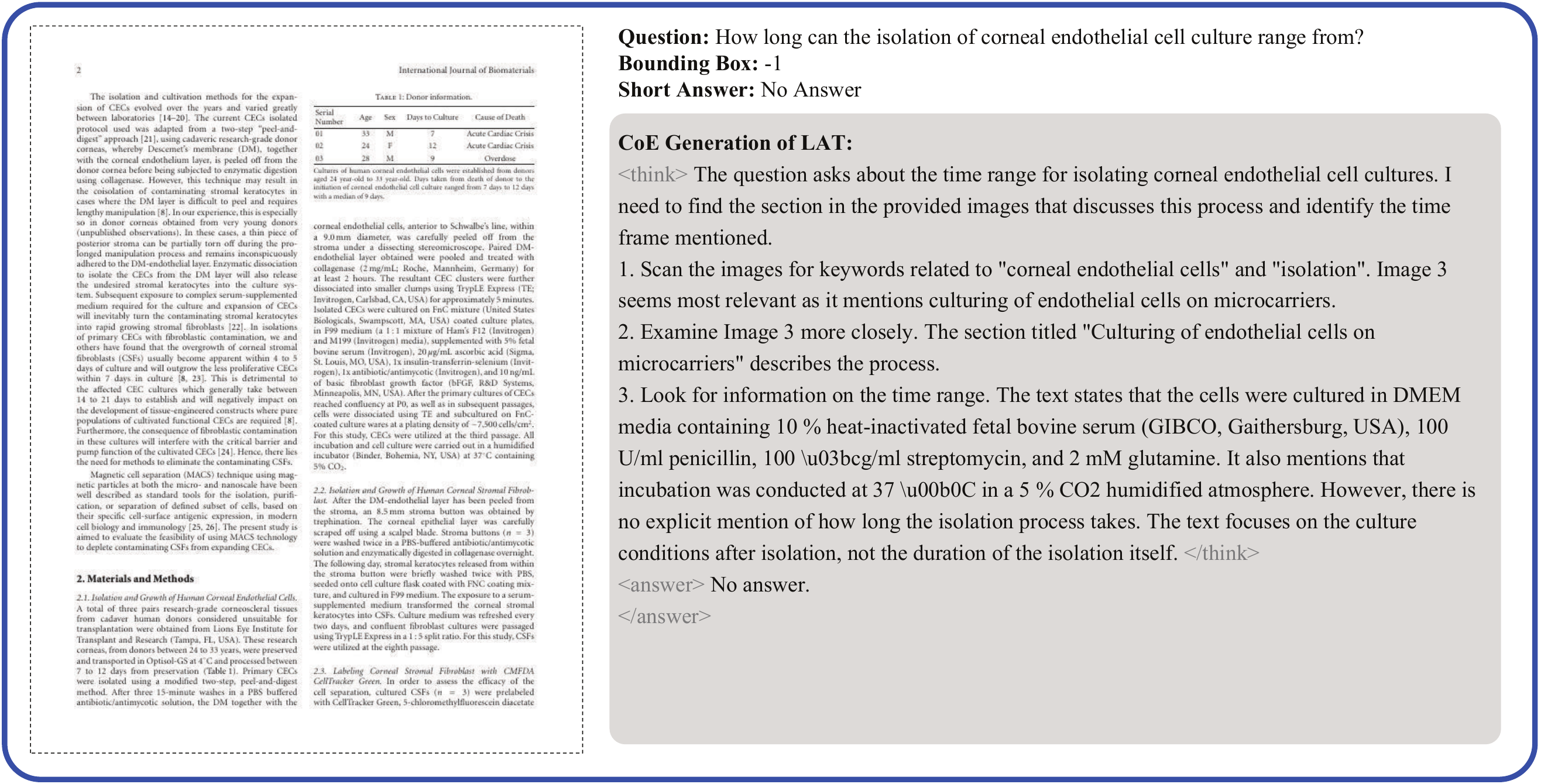} } 
\caption{ 
(Multi-image) The Chain-of-Evidence reasoning process of LAT on the Paper-VISA. This is a case in an unanswerable scenario. (Case 5) 
} 
\label{fig:case5} 
\end{figure*}

\begin{figure*}[t] 
\centerline{
\includegraphics[width=\textwidth]{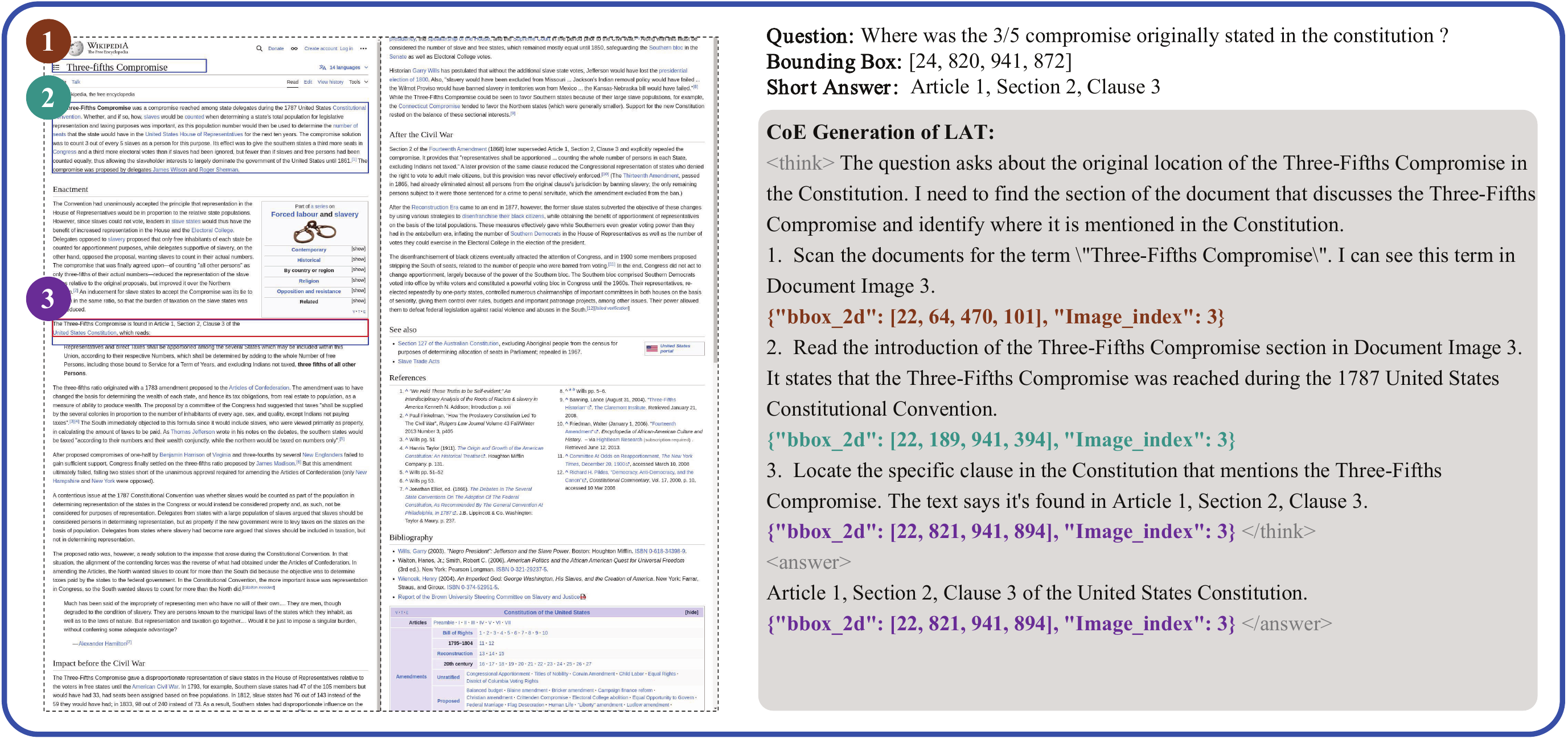} } 
\caption{ 
(Multi-image) The Chain-of-Evidence reasoning process of LAT on the Wiki-VISA. Blue boxes represent model predictions, while the red box indicates the ground truth bounding box from the dataset. (Case 6)
} 
\label{fig:case6} 
\end{figure*}

Figure~\ref{fig:case1}–\ref{fig:case6} present representative examples of model responses. Overall, LAT improves the traceability of the reasoning process. Based on the evaluation of several model responses, we find that LAT generates coherent Chain-of-Evidence (CoE) reasoning traces while maintaining general QA performance and attribution precision. The model is guided by the prompt and CoE training data to directly locate content relevant to the query, following a coarse-to-fine observation process as shown in Figures~\ref{fig:case1} and~\ref{fig:case2}. In Wiki-VISA, where reasoning requires searching across different layouts, CoE enables the model to verify answer correctness by progressively narrowing down evidence. Figures~\ref{fig:case5} and~\ref{fig:case6} show the results under multi-image scenarios. Table~\ref{tab:general3} reports the accuracy of correctly detecting no-answer cases in multi-image settings.

\section{Limitation and Future Work}
While LAT substantially improves stepwise attribution and overall performance, several practical aspects remain open for future enhancement. First, the thresholds used in the stepwise attribution rewards are manually set based on annotated examples, a choice that offers training stability. Exploring adaptive thresholding mechanisms, such as confidence-aware scaling, may improve flexibility and enhance alignment robustness. Second, LAT is currently trained only on the VISA dataset, whose evidence annotations predominantly follow a single-hop structure. As a result, LAT’s ability to generalize to multi-hop or cross-source evidence reasoning is not yet fully evaluated or supervised. Incorporating datasets with explicit multi-hop evidence chains would provide the necessary supervision to assess and strengthen LAT’s capacity for longer reasoning trajectories and cross-page evidence composition.

\end{document}